\documentclass[11pt]{article}

\usepackage[final]{acl}

\usepackage{times}
\usepackage{latexsym}
\usepackage{amsmath}
\usepackage{minted}
\usepackage{subcaption}
\usepackage{float}            
\usepackage{kotex}
\usepackage{tcolorbox}
\usepackage{enumitem}
\usepackage{multirow}
\usepackage{makecell}
\usepackage{pifont}
\usepackage{amssymb}
\usepackage{xcolor}     
\usepackage{tcolorbox}  
\usepackage{amssymb}    
\usepackage{tabularx}   
\usepackage{booktabs}

\usepackage[T1]{fontenc}

\usepackage[utf8]{inputenc}

\usepackage{microtype}

\usepackage{inconsolata}

\usepackage{graphicx}

\title{Progressive Multimodal Search and Reasoning\\for Knowledge-Intensive Visual Question Answering}

\author{
  Changin Choi\textsuperscript{1,3} \quad
  Wonseok Lee\textsuperscript{1} \quad
  Jungmin Ko\textsuperscript{1} \quad
  Wonjong Rhee\textsuperscript{1,2} \\
  \textsuperscript{1}Interdisciplinary Program in Artificial Intelligence, Seoul National University \\
  \textsuperscript{2}Department of Intelligence and Information, Seoul National University \\
  \textsuperscript{3}Samsung Advanced Institute of Technology, Samsung Electronics Co., Ltd \\
  \texttt{\{ci2015.choi, dnjstjr1017, jungminko, wrhee\}@snu.ac.kr}
}

\begin{document}
\maketitle

\maketitle
\begin{abstract}

Knowledge-intensive visual question answering~(VQA) requires external knowledge beyond image content, demanding precise visual grounding and coherent integration of visual and textual information. 
Although multimodal retrieval-augmented generation has achieved notable advances by incorporating external knowledge bases, existing approaches largely adopt single-pass frameworks that often fail to acquire sufficient knowledge and lack mechanisms to revise misdirected reasoning. 
We propose PMSR~(Progressive Multimodal Search and Reasoning), a framework that progressively constructs a structured reasoning trajectory to enhance both knowledge acquisition and synthesis. 
PMSR uses dual-scope queries conditioned on both the latest record and the trajectory to retrieve diverse knowledge from heterogeneous knowledge bases. The retrieved evidence is then synthesized into compact records via compositional reasoning.
This design facilitates controlled iterative refinement, which supports more stable reasoning trajectories with reduced error propagation.
Extensive experiments across six diverse benchmarks~(Encyclopedic-VQA, InfoSeek, MMSearch, LiveVQA, FVQA, and OK-VQA) demonstrate that PMSR consistently improves both retrieval recall and end-to-end answer accuracy.
\end{abstract}

\section{Introduction}
\label{sec:intro}


The emergence of multimodal large language models~(MLLMs) has driven significant progress in multimodal understanding and reasoning.
Nonetheless, recent models continue to struggle with knowledge-intensive visual question answering~(VQA) tasks, which require external knowledge beyond the visual content in the image. These questions require a tightly coupled process of (1) grounding visual entities, (2) retrieving relevant external knowledge, and (3) synthesizing visual and textual evidence to produce an answer.

Multimodal Retrieval-Augmented Generation (RAG) has become a natural solution to this challenge. In the standard RAG process, the model retrieves image–text pairs from an external knowledge base given the input image and question, and then conditions the MLLM on the retrieved context to generate an answer. Recent work has strengthened RAG via improved multimodal retrievers, hierarchical filtering, and reranking~\cite{cocchi2024augmenting, zhang2024mr, ling2025mmkb, chen2024mllm, liu2024lamra, yan-xie-2024-echosight, yang2025omgm}. 

However, this \emph{retrieve-then-read} process is problematic for knowledge-intensive VQA, where initial retrieval is often insufficient, as imperfect retrievers frequently fail to gather all necessary knowledge or introduce distracting passages~\cite{zhang2023siren, shi2023large, Cuconasu2024ThePO, yoran2024making}. 
These limitations are further amplified in multimodal RAG, where distractors in both modalities can mislead reasoning and degrade performance. Textual distractors dominate the model’s attention and bias it toward irrelevant passages, whereas visual distractors can corrupt visual grounding and misdirect reasoning~\cite{deng2025words,bae2025reasoning}. 

Motivated by these limitations, an emerging line of work has explored agentic approaches that leverage reasoning for iterative, tool-augmented retrieval~\cite{li2024omnisearch, geng2025webwatcher, wu2025mmsearch, hong2025deepeyesv2agenticmultimodalmodel}.
In these frameworks, agents reason and act iteratively, conditioning each action on the accumulated interaction history, including prior reasoning traces and tool outputs.
However, errors in query generation, filtering information, and evidence summarization frequently accumulate in these multi-round interactions~\cite {jiang2024mmsearch}. 
Since these frameworks condition each step on the full interaction history, they primarily rely on context accumulation, retaining intermediate reasoning and tool outputs in an ever-growing context. As a result, early errors can propagate through the unstructured history and gradually drift subsequent retrieval and reasoning.


We propose PMSR (Progressive Multimodal Search and Reasoning), which progressively constructs a structured reasoning trajectory to enhance knowledge acquisition and synthesis. Unlike prior approaches that condition each step on the full interaction history, PMSR maintains the reasoning state as a trajectory of compact records synthesized from retrieved evidence, and leverages this trajectory to guide subsequent retrieval and reasoning. Specifically, PMSR is built on two key ideas: \emph{record-isolated updates}, where each iteration synthesizes a new reasoning record solely from newly retrieved evidence, and \emph{dual-scope querying}, which decouples the latest reasoning state from the overall trajectory to support both local retrieval refinement and trajectory-level reflection. To acquire diverse knowledge, PMSR formulates dual-scope queries to retrieve complementary evidence from heterogeneous knowledge bases (KBs). The retrieved evidence from diverse sources is synthesized through compositional reasoning into a compact reasoning record and appended to the trajectory for the next iteration.

We conduct extensive experiments on six knowledge-intensive VQA benchmarks, including Encyclopedic-VQA (E-VQA), InfoSeek, MMSearch, LiveVQA, FVQA, and OK-VQA.
Experimental results demonstrate that PMSR consistently improves retrieval recall and end-to-end answer accuracy over multimodal baselines across various benchmarks, achieving outstanding performance on five benchmarks. Our ablations confirm that the components work synergistically, and trajectory analysis shows that PMSR more often corrects early failures and reduces drift across iterations.

\section{Related work}
\label{sec:related_work}
\begin{figure*}[t]
\begin{center}
   \includegraphics[width=0.9\linewidth]{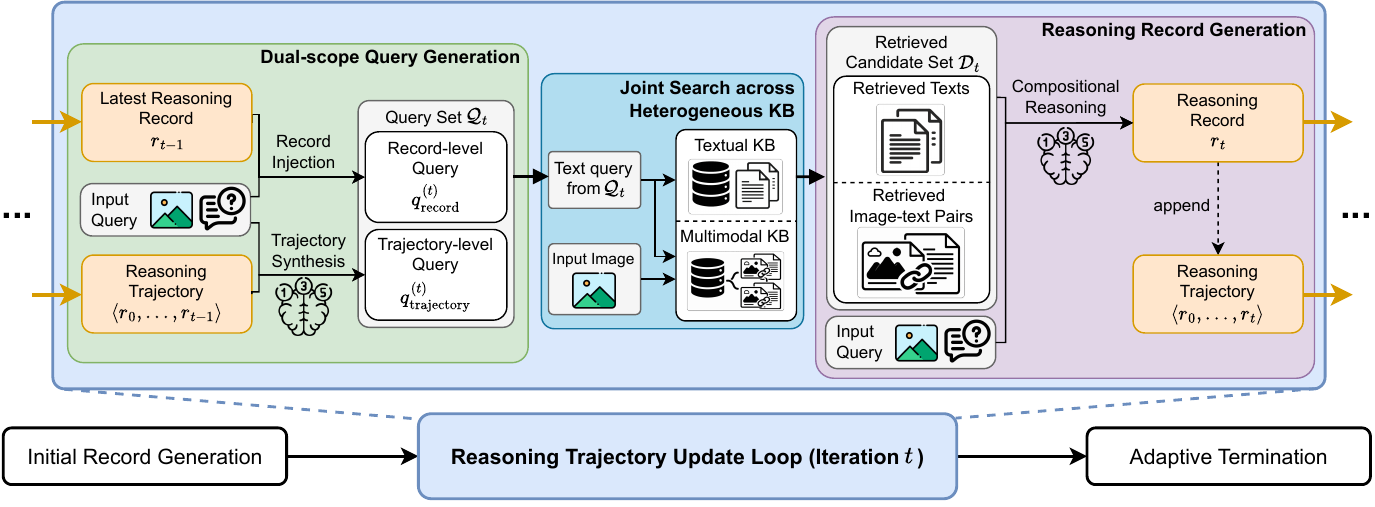}   
    \vspace{-2mm}
\end{center}
\caption{
Overview of PMSR with the reasoning trajectory update loop at iteration $t$.
PMSR consists of three stages: initial record generation, iterative reasoning trajectory updates, and adaptive termination. At each iteration, the reasoning trajectory update loop generates dual-scope queries conditioned on the latest reasoning record and the trajectory, retrieves knowledge from heterogeneous textual and multimodal KBs, and synthesizes the retrieved candidates into a new reasoning record. The newly generated record is appended to the trajectory to guide subsequent iterations.
The process terminates adaptively when further iterations provide limited additional evidence.
}
\vspace{-4mm}
\label{fig:method_overview}
\end{figure*}

\subsection{Multimodal RAG}
Multimodal retrieval-augmented generation (RAG) for knowledge-intensive VQA has largely followed a \emph{retrieve-then-read} paradigm: external knowledge is retrieved in single-step and then read by the model to answer the question. Early work primarily focused on improving single-step retrieval by learning more effective multimodal embeddings~\cite{wei2024uniir,lin-etal-2024-preflmr,lin2025mmembed,liu2024lamra,jiang2025vlmvec,jiang2024e5}. Subsequent approaches extended this paradigm by incorporating coarse-to-fine retrieval strategies. 
Hierarchical systems such as Wiki-LLaVA~\cite{caffagni2024wiki} and EchoSight~\cite{yan-xie-2024-echosight} adopt coarse-to-fine retrieval pipelines with image-based retrieval followed by multimodal or text-based reranking, while OMGM~\cite{yang2025omgm} further develops this paradigm through a multi-step pipeline that explicitly models multiple knowledge granularities via successive multimodal and textual reranking.

More recent studies have focused on enhancing the \emph{read} phase by leveraging the reasoning capabilities of MLLMs. For instance, ReflectiVA~\cite{cocchi2024augmenting} and mR$^2$AG$\dagger$~\cite{zhang2024mr} use self-reflection to evaluate retrieval adequacy and evidence relevance. MMKB-RAG~\cite{ling2025mmkb} generates semantic tags to filter irrelevant evidence. Wiki-PRF~\cite{hong2025wikiprf} adopts reinforcement learning to retain only relevant information.
Despite these advances, the majority of multimodal RAG approaches remain a static \emph{retrieve-then-read} paradigm, constraining the model's ability to refine retrieval as reasoning evolves.

\subsection{Multimodal Agents}
\label{subsec:related_work_agents}

The emergence of agentic paradigms has shifted research from \emph{retrieve-then-read} to agent-based frameworks, where an agent iteratively combines step-by-step reasoning with actions, enabling it to solve complex problems by interacting with external tools~\cite{yao2022react}. Early explorations of this direction appeared in \emph{iterative RAG} for text-only question answering~\cite{wang2024rat, xiong2024improving, yue2025inference, trivedi2023interleaving, yu2024auto_rag, jiang2024retrieve_summarize, zhang2025levelrag, Liu-RA-ISF}, where models decompose complex queries into sub-queries and iteratively perform retrieval within predefined workflows.

Building on these ideas, multimodal agents bring retrieval and tool use to vision–language settings by coupling retrieval with tool-augmented interaction, enabling models to iteratively reformulate queries and select external tools during multi-step reasoning. OmniSearch~\cite{li2024omnisearch} introduces adaptive planning that routes multimodal queries across multiple search tools. More recent work learns search and tool-use policies via reinforcement learning: WebWatcher~\cite{geng2025webwatcher} and MMSearch-R1~\cite{wu2025mmsearch} optimize retrieval trajectories over tool interactions, internalizing decision-making across steps. DeepEyesV2~\cite{hong2025deepeyesv2agenticmultimodalmodel} further integrates perception, search, and code execution within agentic loop.

However, these agentic approaches condition each step on a long interaction history, such that earlier intermediate outputs remain in the conditioning and continue to influence subsequent retrieval and reasoning. 
In contrast, our method progressively retrieves evidence using dual-scope queries over heterogeneous KBs and condenses retrieved knowledge as a record to update the reasoning trajectory, mitigating drift from earlier intermediate outputs.

\section{Method}
\label{sec:method}

PMSR~(Progressive Multimodal Search and Reasoning) is a framework for knowledge-intensive VQA where it progressively constructs a reasoning trajectory, as illustrated in Figure~\ref{fig:method_overview}.


\subsection{Initial Reasoning Record Generation}
\label{sec:initial_record_generation}

To bootstrap the reasoning trajectory, PMSR first constructs an initial reasoning record using an MLLM. 
Unlike later iterations that build upon the trajectory, this step combines the model’s parametric knowledge with externally retrieved knowledge.

Given the input query composed of an image $I$ and a question $Q$, the MLLM generates a visually grounded description relevant to the question:
\begin{equation}
    d_0 = \mathcal{G}_{\text{desc}}(Q, I).
\end{equation}

We then expand the query by concatenating $Q$ with this description, forming an enriched query $q_{\text{init}} = [Q; d_0]$. 
Using $q_{\text{init}}$, PMSR retrieves an initial candidate set $\mathcal{D}_0$ from the heterogeneous KBs.

Finally, the retrieved candidate set is synthesized into the first reasoning record using a dedicated reasoning operator:
\begin{equation}
    r_0 = \mathcal{G}_{\text{reason}}(Q, I, \mathcal{D}_0).
\end{equation}
This produces a coherent summary of the relevant facts, initializing the reasoning trajectory $\langle r_0\rangle$.

\subsection{Dual-scope Query Formulation}
\label{sec:multi_query_formulation}
After initialization, PMSR progressively guides knowledge search by generating new queries conditioned on the evolving reasoning trajectory. PMSR decomposes query generation into two complementary scopes: a record-level query grounded in the latest reasoning record and a trajectory-level query derived from the accumulated reasoning trajectory. The record-level query supports local refinement by using the latest reasoning record to retrieve evidence closely related to the most recent deduction, whereas the trajectory-level query supports global reflection by analyzing compact records in the trajectory to identify unresolved gaps, resolve conflicts, and retrieve broader contextual evidence.

The set of queries generated at iteration $t$ is given by:
\begin{equation}
    \mathcal{Q}_t
    = \left\{ q_{\text{record}}^{(t)},\; q_{\text{trajectory}}^{(t)} \right\}.
\end{equation}

\vspace{1mm}
\noindent\textbf{Record-level query.}
The record-level query conditions on the latest reasoning record $r_{t-1}$, using its most recent deductions to expand the query to retrieve additional knowledge relevant to the current reasoning state. In the standard PMSR setting for knowledge-intensive VQA, this is implemented by concatenating the input question with $r_{t-1}$:
\begin{equation}
    q_{\text{record}}^{(t)} = [\, Q;\; r_{t-1} \,].
\end{equation}
For the web-equipped variant of PMSR, this operator is adapted to produce a compact reformulation suitable for search engine constraints.

\vspace{1mm}
\noindent\textbf{Trajectory-level query.}
The trajectory-level query leverages the reasoning trajectory to retrieve knowledge guided by the evolving reasoning records. Formally, a dedicated operator synthesizes information from this trajectory to generate a context-specific query:
\begin{equation}
    q_{\text{trajectory}}^{(t)} =
    \mathcal{G}_{\text{trajectory}}\left(Q, I, \langle r_0, \dots, r_{t-1} \rangle \right).
\end{equation}
In contrast to the record-level query, it incorporates broader context accumulated over prior reasoning steps, where its goal is to guide the proper direction for the next query given reasoning records.


\subsection{Joint Search across Heterogeneous KBs}
\label{sec:heterogenous_kb}
To support compositional reasoning with diverse external knowledge, PMSR performs a joint search over heterogeneous KBs using the dual-scope query set generated at each iteration. Given the query set $\mathcal{Q}_t$, PMSR retrieves candidates from a textual KB and a multimodal KB. 

\vspace{1mm}
\noindent \textbf{Retrieval from textual KB.}
For each query $q_t \in \mathcal{Q}_t$, we retrieve passages $p$ from the textual KB using text-text semantic similarity:
\begin{equation}
    S_{\text{txt}} = \operatorname{sim}_{\text{text}}(q_t, p),
\end{equation}
where $\operatorname{sim}_{\text{text}}$ denotes cosine similarity in a text embedding space.

\vspace{1mm}
\noindent \textbf{Retrieval from multimodal KB.}
The multimodal KB consists of image-text pairs~$(I_c, t_c)$. For each query $q_t \in \mathcal{Q}_t$ and the input image $I$, we compute a decoupled similarity score:
\begin{equation}
\begin{split}
    S_{\text{mm}}
    = \lambda\, \operatorname{sim}_{\text{text}}(q_t, t_c)
    + (1-\lambda)\, \operatorname{sim}_{\text{img}}(I, I_c),
\end{split}
\end{equation}
where $\operatorname{sim}_{\text{img}}$ denotes cosine similarity in an image embedding space, and we use a fixed weight of $\lambda = 0.5$ to balance the two modalities. The text term adapts retrieval to the dual-scope query, while the image term preserves visual relevance to the input image.


\vspace{1mm}
\noindent \textbf{Combined retrieval.}
We retrieve up to $N_{\text{txt}}{=}20$ text passages and $N_{\text{mm}}{=}10$ image-text pairs per iteration and aggregate them into the candidate set $\mathcal{D}_t$. We evenly split the retrieval budget between record- and trajectory-level queries. Additional implementation details are provided in Appendix~\ref{sec:multimodal_retrieval}.

\subsection{Reasoning Record Generation}
\label{sec:reasoning_record_generation}
After retrieving candidates from heterogeneous KBs, PMSR constructs a reasoning record from the newly retrieved knowledge.

At iteration $t$, the retrieved candidate set $\mathcal{D}_t$ is synthesized using a dedicated reasoning operator:
\begin{equation}
    r_t = \mathcal{G}_{\text{reason}}(Q, I, \mathcal{D}_t).
\end{equation}
The operator $\mathcal{G}_{\text{reason}}$ integrates retrieved visual and textual knowledge conditioned on the input query, producing a reasoning record. PMSR supports compositional reasoning by aggregating diverse knowledge from heterogeneous KBs into records to guide subsequent iterations.

Importantly, each reasoning record $r_t$ is generated solely from the newly retrieved candidate set $\mathcal{D}_t$, without directly conditioning on previous reasoning records. The resulting record is appended to the reasoning trajectory, yielding $\langle r_0, \dots, r_{t-1}, r_t \rangle$ for subsequent iterations.


\subsection{Adaptive Termination via Information Saturation}
\label{sec:adaptive_termination}
To improve inference efficiency, we introduce an adaptive termination criterion based on information saturation, where increasing similarity between newly generated and earlier queries indicates redundant retrieval. 
This similarity is quantified by the saturation score, defined as
\begin{equation}
    \delta_{\text{query}}^{(t)} =
    \max_{q \in \mathcal{Q}_t,\, q' \in \mathcal{Q}_j,\, j < t}
    \operatorname{sim}_{\text{text}}(q, q').
\end{equation}
The iterative process terminates when
\begin{equation}
    \delta_{\text{query}}^{(t)} \ge \tau.
\end{equation}
Unless otherwise stated, we set $\tau = 0.9$ in all experiments. Upon termination at iteration $T$, the MLLM generates the final answer conditioned on $Q$, $I$, and the reasoning trajectory $\langle r_0, \dots, r_T \rangle$.

The prompt templates used to instantiate the MLLM operators
($\mathcal{G}_{\text{desc}}$, $\mathcal{G}_{\text{trajectory}}$, and $\mathcal{G}_{\text{reason}}$)
are provided in Appendix~\ref{sec:prompts_appendix}. For the web-equipped variant of PMSR, the prompt and details of implementations are provided in Appendix~\ref{sec:web_prompts} and ~\ref{sec:web_search}.


%
%

\section{Experiments}
\label{sec:experiments}
To evaluate the performance of our proposed PMSR framework, we conduct experiments on several challenging benchmark datasets using a diverse set of evaluation metrics.

\begin{table*}[!t]
\centering
\small
\begin{tabular}{lcccccc}
\toprule
\textbf{Method} & \multicolumn{3}{c}{\textbf{InfoSeek}} & \multicolumn{3}{c}{\textbf{E-VQA}} \\
\cmidrule(lr){2-4} \cmidrule(lr){5-7}
& R@5 & R@10 & R@20 & R@5 & R@10 & R@20 \\
\midrule
Wiki-LLaVA~\cite{caffagni2024wiki} & -     & 66.1 & 71.9 & -    & \hphantom{0}9.9  & 13.2  \\
LLM-RA~\cite{jian2024large}        & 53.8  & -    & -    & -    & -    & -     \\
mR²AG~\cite{zhang2024mr}           & -     & 65.0 & 71.0 & -    & -    & -     \\
ReflectiVA~\cite{cocchi2024augmenting} & \underline{77.6} & - & \textbf{86.4} & 36.1 & - & \underline{49.8} \\
EchoSight$\dagger$~\cite{yan-xie-2024-echosight} & 74.0 & 77.4 & 77.9 & \underline{47.9} & 48.8 & 48.8 \\
OMGM~\cite{yang2025omgm}           & 73.9 & \underline{80.0} & 84.8 & 41.2 & \underline{49.8} & \textbf{58.7} \\
OMGM$\dagger$~\cite{yang2025omgm}  & \textbf{80.8} & \textbf{83.6} & \underline{84.8} & \textbf{55.7} & \textbf{58.1} & \textbf{58.7} \\
ReAuSE~\cite{long2025retrieval}    & 59.5 & -     & -    & -    & -    & -     \\
\midrule
\multicolumn{1}{l}{} & \multicolumn{6}{c}{\textit{Cumulative Recall}} \\
\midrule
Ours (Qwen3-VL-4B)* & \multicolumn{3}{c}{\underline{87.4}} & \multicolumn{3}{c}{\underline{64.3}} \\
Ours (Qwen3-VL-8B)* & \multicolumn{3}{c}{\textbf{87.7}} & \multicolumn{3}{c}{\textbf{67.3}} \\
\bottomrule
\end{tabular}
\caption{Recall comparison on the InfoSeek validation and E-VQA test sets.
$\dagger$ indicates methods that utilize reranking.
For PMSR (*), we report cumulative recall at adaptive termination.
Best and second-best results are highlighted in bold and underlined, respectively.}
\vspace{-4mm}
\label{tab:Recall_InfoSeek_E-VQA}
\end{table*}

\subsection{Experiment Setup}
\paragraph{Datasets.}
We evaluate PMSR on an extensive suite of knowledge-intensive VQA benchmarks covering encyclopedic, factual, and real-world information-seeking scenarios.
Our experiments use the InfoSeek validation split of M2KR~\cite{lin-etal-2024-preflmr}, the OK-VQA validation split, and the single-hop questions of the E-VQA test split~\cite{mensink2023encyclopedic,chen-etal-2023-infoseek,marino2019ok}, following standard practice.
Moreover, we extend our evaluation of PMSR to four search-oriented benchmarks: FVQA test split, the InfoSeek Human subset, LiveVQA, and MMSearch~\cite{jiang2024mmsearch,wu2025mmsearch,fu2025seekingupdatinglivevisual}. These benchmarks target real-world questions requiring factual grounding, time-sensitive news, and long-tail knowledge.
The details of each benchmark are provided in Appendix~\ref{sec:dataset}.

\paragraph{Knowledge bases and retrievers.}
For a fair comparison, we use fixed heterogeneous KBs across all experiments. 
The multimodal KB consists of 2M Wikipedia image-text pairs provided in InfoSeek, while the textual KB comprises approximately 21M Wikipedia passages from FlashRAG~\cite{jin2024flashrag}. 

For retrieval, we adopt dense similarity search.
Multimodal retrieval uses SigLIP2~\cite{tschannen2025siglip} for image embeddings and Qwen3-Embedding~\cite{zhang2025qwen3embedding} for text embeddings, while textual retrieval uses E5-base-v2~\cite{wang2022text}.
Comparisons with retriever baselines are provided in Appendix~\ref{sec:multimodal_retrieval}.

\paragraph{Multimodal large language models.}
To assess how performance scales with reasoning capacity while ensuring fair comparison, we evaluate two tiers of MLLM backbones: open-source models from the Qwen-VL series (Qwen2.5-VL~\cite{bai2025qwen25vltechnicalreport}, Qwen3-VL~\cite{bai2025qwen3vltechnicalreport}) and the proprietary Gemini-2.5-Flash~\cite{comanici2025gemini}.
\paragraph{Evaluation metrics.}
We evaluate PMSR using standard accuracy and retrieval metrics across benchmarks. For accuracy, we report the official BERT matching score~(BEM)~\cite{bulian-etal-2022-tomayto} on E-VQA and exact match~(EM) on InfoSeek. We additionally report cover exact match~(CEM)~\cite{jiang2024rag,yue2025inference} as a complementary metric that checks whether the ground-truth answer appears in the model output. For OK-VQA, FVQA-test, InfoSeek Human, MMSearch, and LiveVQA, we adopt an LLM-as-a-Judge protocol following MMSearch-R1~\cite{wu2025mmsearch} using GPT-4o; the evaluation prompts are provided in Appendix~\ref{sec:prompt_for_judge}.

For retrieval performance, we measure recall based on the presence of ground-truth evidence in the retrieved context.
We report entity recall for InfoSeek and E-VQA, and Pseudo-Relevance Recall~(PRR)~\cite{luo-etal-2021-prr} for OK-VQA following PreFLMR~\cite{lin-etal-2024-preflmr}. To assess progressive knowledge acquisition, we further report cumulative recall under adaptive termination. All reported results are obtained from a single evaluation run for each model and benchmark.

\begin{table}[!t]
\centering
\small
\resizebox{\columnwidth}{!}
{
\begin{tabular}{lcc}
\toprule
\textbf{Method} & \multicolumn{2}{c}{\textbf{OK-VQA}} \\
\cmidrule(lr){2-3}
& PRR@5 & PRR@10 \\
\midrule
DPR~\cite{karpukhin2020dense} & 66.9 & 76.4 \\
ReViz-ICT~\cite{luo2023end} & 61.9 & 72.6 \\
GeMKR~\cite{long2024generative} & 70.8 & 79.1 \\
FLMR~\cite{lin2023finegrained} & 68.1 & 78.0 \\
Pre-FLMR~\cite{lin-etal-2024-preflmr} & 68.6 & - \\
ReAuSE~\cite{long2025retrieval} & \textbf{88.0} & \textbf{91.3} \\
OMGM$\dagger$~\cite{yang2025omgm} & 73.4 & - \\
\midrule
\multicolumn{1}{l}{} & \multicolumn{2}{c}{\textit{Cumulative Recall}} \\
\midrule
Ours (Qwen3-VL-4B)* & \multicolumn{2}{c}{92.1} \\
Ours (Qwen3-VL-8B)* & \multicolumn{2}{c}{\textbf{97.1}} \\
\bottomrule
\end{tabular}
}
\caption{Recall comparison on the OK-VQA benchmark using Wikipedia as the knowledge source. $\dagger$ indicates methods that utilize reranking.
For PMSR (*), we report the cumulative recall at adaptive termination.}
\label{tab:Recall_OK-VQA}
\vspace{-4.0mm}
\end{table}

\subsection{Retrieval Performance on VQA Benchmarks}

Table~\ref{tab:Recall_InfoSeek_E-VQA} reports the retrieval performance of PMSR on the InfoSeek and E-VQA benchmarks.
Across both datasets, PMSR achieves consistent and substantial improvements over prior methods.

On InfoSeek, the 4B model reaches a cumulative recall of 87.4\%, which is slightly higher than the previous best result reported by ReflectiVA~(86.4\% at R@20). On E-VQA, the 4B model achieves 64.3\% cumulative recall, exceeding OMGM$\dagger$ by 5.6 percentage points.
Scaling the backbone from 4B to 8B further yields consistent gains, improving cumulative recall to 87.7\% on InfoSeek and 67.3\% on E-VQA.

We report cumulative recall for PMSR to reflect progressive evidence accumulation under adaptive stopping. For completeness, we also report per-iteration recall and contributions of each KB in Appendix~\ref{sec:hetero_kb_iteration}. To quantify the efficiency gains enabled by adaptive stopping, Section~\ref{subsec:ablation_adaptive} presents an ablation study. 

As shown in Table~\ref{tab:Recall_OK-VQA}, PMSR demonstrates strong and consistent retrieval performance on the OK-VQA benchmark, despite highly different from other benchmarks in knowledge type, question formulation, and grounding requirements.
On OK-VQA, PMSR achieves 92.1\% and 97.1\% cumulative recall with the 4B and 8B models, respectively, demonstrating strong retrieval performance across knowledge types.
This cross-domain stability contrasts sharply with prior retrieval-augmented models, which often perform well only within their target domain. The results indicate that PMSR’s progressive retrieval strategy generalizes effectively across knowledge types without requiring dataset-specific tuning. 

Additionally, Appendix~\ref{sec:OKVQA_Accuracy} reports end-to-end answer accuracy under an LLM-as-a-Judge protocol. Appendix~\ref{sec:record_ablation} further analyzes the contribution of the latest reasoning record, while Appendix~\ref{sec:lambda_ablation} provides a sensitivity study on the interpolation weight λ.

\vspace{-1.0mm}
\subsection{Accuracy on Search Benchmarks}
\vspace{-1.0mm}

\begin{table}[!t]
\centering
\small
\resizebox{1.0\columnwidth}{!}{
\begin{tabular}{l c c c c}
\toprule
\textbf{Method} & \textbf{\makecell{FVQA\\test}} & \textbf{\makecell{InfoSeek\\Human}} & \textbf{\makecell{MM\\Search}} & \textbf{\makecell{Live\\VQA}} \\
\midrule
\makecell[l]{OmniSearch (GPT-4o)\\~\cite{li2024omnisearch}} & - & -  & 49.7  & 40.9 \\
\makecell[l]{MMSearch-R1\\~\cite{wu2025mmsearch}} & 58.4 & 55.1 & 53.8 & 48.4 \\
\makecell[l]{WebWatcher\\~\cite{geng2025webwatcher}} & - & -  & 49.1  & 51.2 \\
\makecell[l]{DeepEyesV2\\~\cite{hong2025deepeyesv2agenticmultimodalmodel}} & 60.6 & 51.1 & \textbf{63.7} & - \\
\midrule
Ours & \textbf{61.2} & \textbf{58.2} & 54.3 & \textbf{54.2} \\
\bottomrule
\end{tabular}
}
\caption{Performance on search-oriented multimodal benchmarks. Results for OmniSearch are taken from WebWatcher~\cite{geng2025webwatcher}. Unless otherwise noted, all methods use Qwen2.5-VL-7B as the backbone, ensuring a fair comparison.}
\vspace{-7mm}
\label{tab:search_benchmarks}
\end{table}

We evaluate PMSR on search-oriented benchmarks that require multimodal grounding and open-domain knowledge acquisition. As shown in Table~\ref{tab:search_benchmarks}, using the same Qwen2.5-VL-7B backbone, PMSR achieves 61.2\% and 58.2\% accuracy on FVQA and InfoSeek Human, respectively, surpassing recent agent-based baselines. On MMSearch, PMSR attains 54.3\% accuracy, remaining competitive with specialized multimodal search agents. On LiveVQA, which targets real-world, time-sensitive information seeking over diverse news sources, PMSR reaches 54.2\% accuracy, the highest among the methods reported in Table~\ref{tab:search_benchmarks}.

\begin{table*}[h]
\centering
\small

\resizebox{\textwidth}{!}{
\begin{tabular}{l c c c c c}
\toprule
& & & \multicolumn{2}{c}{\textbf{InfoSeek}} & \textbf{E-VQA} \\
\cmidrule(lr){4-5} \cmidrule(lr){6-6}
\textbf{Method} & \textbf{Retriever} & \textbf{Model} & \textbf{Val} & \textbf{M2KR} & \textbf{Single-hop} \\
\midrule
Wiki-LLaVA~\cite{caffagni2024wiki} & CLIP-ViT-L & LLaVA-1.5-7B & 28.9 & - & 21.8 \\
EchoSight$\dagger$~\cite{yan-xie-2024-echosight} & EVA-CLIP-8B & Mistral-7B & 31.3 & - & 35.5 \\
LLM-RA~\cite{jian2024large} & EVA-CLIP-8B & BLIP2-Flan-T5XL & 23.1 & - & - \\
mR²AG$\dagger$~\cite{zhang2024mr} & CLIP-ViT-L & LLaVA-1.5-7B & 40.2 & - & - \\
ReflectiVA~\cite{cocchi2024augmenting} & EVA-CLIP-8B & LLaVA-MORE-8B & 40.1 & - & 35.5 \\
MMKB-RAG$\dagger$~\cite{ling2025mmkb} & PreFLMR ViT-G & Qwen2-VL-7B & 36.7 & 34.7 & 39.7 \\
RET-2 ~\cite{caffagni2025recurrence} & RET-2 & LLaVA-MORE-8B & 22.8 & - & 28.5 \\
Wiki-PRF(w/ RL)~\cite{hong2025wikiprf} & EVA-CLIP-8B & VLM-PRF-7B & \underline{42.5} & - & 40.1 \\
OMGM$\dagger$~\cite{yang2025omgm} & EVA-CLIP-8B & LLaVA-1.5-7B & \textbf{43.5} & - & \underline{50.2} \\
OMGM$\dagger$ & EVA-CLIP-8B & GPT-4o & 42.1 & - & \textbf{51.2} \\
\midrule
\multirow{3}{*}{Ours} & \multirow{3}{*}{SigLIP2-g} & Qwen3-VL-4B & - & 38.3* & 40.9 \\
 & & Qwen3-VL-8B & - & \underline{41.5}* & \underline{46.4} \\
 & & Gemini-2.5-Flash & - & \textbf{50.5}* & \textbf{59.9} \\
\bottomrule
\end{tabular}
}
\caption{Overall accuracy on InfoSeek and E-VQA.
\textbf{Val} denotes the full InfoSeek validation set~(137K), and \textbf{M2KR} the 5K subset.
E-VQA results are reported on the single-hop subset.
$\dagger$ indicates methods that utilize reranking; * indicates that LLaVA-MORE-8B~(ReflectiVA) is used as the final answer generator for EM evaluation.}
\vspace{-4mm}
\label{tab:Merged_Accuracy}
\end{table*}


\subsection{Accuracy on Knowledge-Intensive VQA}
\label{sec:overall_accuracy}

We report end-to-end answer accuracy of PMSR on the InfoSeek and E-VQA benchmarks in Table~\ref{tab:Merged_Accuracy}.
Across both datasets, PMSR achieves strong performance compared to prior retrieval-augmented approaches, highlighting the effectiveness of progressive, reasoning-guided retrieval.

On the E-VQA benchmark, PMSR with Qwen3-VL-8B achieves 46.4\% accuracy, which is comparable to strong prior baselines. When the trajectory is generated using a more capable model~(Gemini-2.5-Flash), accuracy increases to 59.9\%, surpassing the previous best by 8.7\%.

On the InfoSeek M2KR subset, PMSR also demonstrates substantial improvements. Performance scales with the capacity of the reasoning backbone, with the Qwen3-VL-8B configuration achieving 41.5\% accuracy and the Gemini-2.5-Flash configuration reaching 50.5\% accuracy.

Importantly, for InfoSeek, we evaluate accuracy using LLaVA-MORE-8B as a final answerer, regardless of which MLLM is used to generate the reasoning records.
This controlled setup isolates the contribution of PMSR: improvements on InfoSeek reflect reasoning trajectory produced by PMSR, rather than differences in the answer generation model.
Accordingly, stronger trajectory-generation configurations (e.g., Gemini-2.5-Flash) yield higher accuracy because they construct more informative and better-grounded reasoning trajectories, which the same answerer can exploit more effectively. Additional qualitative examples illustrating these trajectories are provided in Appendix~\ref{sec:qualitative_examples}.

\vspace{-1.0mm}
\section{Ablations}
\vspace{-1.0mm}
\label{sec:ablation}

For ablation studies, we conduct experiments using Qwen3-VL-8B to validate the robustness of individual components. 
Unless otherwise specified, the retrieval budget is fixed across single-query and dual-scope query settings, and adaptive termination is used with $\tau = 0.9$ (up to a maximum of 5 iterations). Specifically, under the heterogeneous KB setting, we retrieve a total of 20 text passages and 10 image-text pairs per iteration; for dual-scope querying, this budget is evenly split across the two queries. When using only the multimodal KB, we retrieve 10 image-text pairs per iteration.

\subsection{Impact of Iterative Performance}
\label{subsec:iterative_performance}

\begin{table}[!h]
\centering
\small
\resizebox{\columnwidth}{!}
{
\begin{tabular}{c cc cc} 
\toprule
\multirow{2}{*}{\textbf{Iter.}} & \multicolumn{2}{c}{\textbf{InfoSeek}} & \multicolumn{2}{c}{\textbf{E-VQA}} \\
\cmidrule(lr){2-3} \cmidrule(lr){4-5}
& \textbf{CEM} & \textbf{Recall} & \textbf{BEM} & \textbf{Recall} \\
\midrule
0 & 48.3        & 82.8        & 37.1        & 59.0 \\
1 & 53.6 (+5.3) & 86.0 (+3.2) & 42.2 (+5.1) & 63.8 (+4.8) \\
2 & 54.8 (+1.2) & 87.5 (+1.5) & 45.4 (+3.2) & 65.5 (+1.7) \\
3 & 55.4 (+0.6) & 88.1 (+0.6) & 46.2 (+0.8) & 66.6 (+1.1) \\
4 & 56.3 (+0.9) & 88.5 (+0.4) & 47.1 (+0.9) & 67.5 (+0.9) \\
\bottomrule
\end{tabular}
}
\caption{Performance of PMSR across iterations on InfoSeek and E-VQA. Results are reported for a fixed sequence of 5 iterations; Iteration 0 corresponds to the initial reasoning record.}
\label{tab:ablation_iterative_rag}
\vspace{-4.0mm}
\end{table}


To quantify the effect of progressive search and reasoning, Table~\ref{tab:ablation_iterative_rag} reports performance as the number of iterations increases. Across both InfoSeek and E-VQA, PMSR exhibits monotonic improvements in both accuracy~(CEM/BEM) and retrieval recall. The largest improvements occur in the first one to two iterations, after which improvements become smaller but remain consistent. These results indicate that progressively accumulating reasoning records and using them to guide subsequent retrieval provides measurable benefits. To further examine how intermediate reasoning evolves across iterations, we analyze reasoning trajectory dynamics in Section~\ref{subsec:reasoning_trajectory}.


\subsection{Ablation of components}

\begin{table}[h!]
\centering
\small
\begin{tabular}{ccc cc}
\toprule
\textbf{\makecell{Dual-scope\\Query}} & \textbf{\makecell{Hetero.\\KB}} & \textbf{Iter.} & \textbf{BEM} & \textbf{Recall} \\
\midrule
$\times$ & $\times$ & $\times$ & 30.7 & 48.7 \\
$\checkmark$ & $\times$ & $\times$ & 32.4 & 48.8 \\
$\times$ & $\checkmark$ & $\times$ & 34.4 & 58.4 \\
$\checkmark$ & $\checkmark$ & $\times$ & 37.3 & 58.8 \\
$\checkmark$ & $\times$ & $\checkmark$ & 34.6 & 56.3 \\
$\times$ & $\checkmark$ & $\checkmark$ & 43.1 & 64.8 \\
\midrule
\textbf{\checkmark} & \textbf{\checkmark} & \textbf{\checkmark} & \textbf{46.4} & \textbf{67.3} \\
\bottomrule
\end{tabular}
\caption{Component ablation of PMSR on the E-VQA test set. We ablate dual-scope query formulation, retrieval over heterogeneous KBs, and progressive search and reasoning over iterations. Removing dual-scope query uses only the trajectory-level query; removing heterogeneous KBs restricts retrieval to the multimodal KB; removing iterations corresponds to single-pass RAG.}
\label{tab:ablation_final}
\vspace{-2.0mm}
\end{table}

Table~\ref{tab:ablation_final} examines the contribution of key components of PMSR on E-VQA.
Adding a textual KB alongside the multimodal KB substantially improves retrieval recall (48.7$\rightarrow$58.4) and increases BEM (30.7$\rightarrow$34.4), highlighting the importance of heterogeneous KBs.
Enabling dual-scope query formulation further boosts performance, with larger gains observed under heterogeneous KB retrieval (BEM 34.4$\rightarrow$37.3; recall 58.4$\rightarrow$58.8).
Finally, introducing progressive search and reasoning over iterations yields additional improvements, and the full model achieves the best overall performance.
These results indicate that repeated reasoning-guided retrieval and accumulation of reasoning records provide complementary benefits beyond any single component.

\subsection{Ablation of Adaptive Termination}
\label{subsec:ablation_adaptive}
\begin{table}[h]
    \centering
    \small
    \resizebox{\columnwidth}{!}{%
    \begin{tabular}{l|ccc|ccc}
    \toprule
    \multirow{2}{*}{\textbf{Method}} 
    & \multicolumn{3}{c|}{\textbf{InfoSeek}} 
    & \multicolumn{3}{c}{\textbf{E-VQA}} \\ 
     & \textbf{\makecell{Avg.\\Iter.}} & \textbf{CEM} & \textbf{Recall} 
     & \textbf{\makecell{Avg.\\Iter.}} & \textbf{BEM} & \textbf{Recall} \\ 
    \midrule
    Fixed & 5.0 & 56.3 & 88.5 & 5.0 & 47.1 & 67.5 \\
    Adaptive & 3.3 & 55.1 & 87.7 & 3.5 & 46.4 & 67.3 \\ 
    \bottomrule
    \end{tabular}%
    }
    \caption{Impact of adaptive termination compared with a fixed-iteration strategy~(fixed: 5 iterations vs.\ adaptive: $\tau=0.9$).}
    \label{tab:adaptive_ablation}
\vspace{-2mm}
\end{table}



We evaluate the efficiency of adaptive termination by comparing it against a fixed-iteration strategy with the same backbone~(Qwen3-VL-8B). As shown in Table~\ref{tab:adaptive_ablation}, adaptive termination with the default threshold $\tau=0.9$ reduces the average number of iterations from 5.0 to 3.3 on InfoSeek and 3.5 on E-VQA, while maintaining comparable accuracy and retrieval recall.
\section{Analysis}
\label{sec:analysis}

\subsection{Analysis of Reasoning Trajectory}
\label{subsec:reasoning_trajectory}
\begin{table}[h]
\centering
\small
\begin{tabular}{lcc}
\toprule
\textbf{Trajectory Type} & \textbf{InfoSeek} & \textbf{E-VQA} \\
\midrule
Stable-Correct & 43.95\% & 47.73\% \\
Persistent-Fail & 36.79\% & 4.91\% \\
Correction & 11.64\% & 30.06\% \\
Conflict & 7.63\% & 17.31\% \\
\bottomrule
\end{tabular}
\caption{Distribution of reasoning trajectory types based on per-iteration correctness patterns~(CEM on InfoSeek; BEM on E-VQA): \textit{Stable-Correct}~(correct at all iterations), \textit{Persistent-Fail}~(incorrect at all iterations), \textit{Correction}~(recovers from incorrect reasoning and is correct at the final iteration), and \textit{Conflict}~(correct in some iterations but incorrect at the final iteration).}
\label{tab:reasoning_record_stats}
\vspace{-3.0mm}
\end{table}




We analyze reasoning trajectories by evaluating the correctness of each intermediate reasoning record across iterations, rather than only the final prediction, to characterize how reasoning evolves. 
As shown in Table~\ref{tab:reasoning_record_stats}, \textit{Correction} occurs more frequently than \textit{Conflict}, suggesting that the proposed framework more often recovers from early incorrect reasoning than propagates it to later iterations.
Moreover, a substantial portion of trajectories is categorized as \textit{Stable-Correct}, indicating that once correct grounding and relevant knowledge are established, the reasoning process tends to preserve correctness across subsequent iterations.

Notably, on E-VQA, the combined proportion of \textit{Stable-Correct} and \textit{Correction} trajectories exceeds 70\%, indicating that reasoning records progressively gather sufficient knowledge to address the question. However, this proportion is considerably higher than the final answer accuracy, revealing a gap between the quality of intermediate reasoning records and the model’s ability to fully utilize them for answer prediction. To further examine this gap, we present a model sensitivity analysis of the contextual noise of distractors in Appendix~\ref{sec:contextual_noise}. Furthermore, we provide a trajectory type comparison with the web search agent in Appendix~\ref{sec:traj_comp_webwatcher} under the same KB and retriever.

\vspace{-1.0mm}
\subsection{Analysis of Adaptive Termination}
\vspace{-1.0mm}
\label{appendix:adaptive_convergence}
\begin{table}[h]
\centering
\small
\begin{tabular}{lcc}
\toprule
\textbf{Trajectory Category} & \textbf{InfoSeek} & \textbf{E-VQA} \\
\midrule
Stable-Correct & 1.96 & 2.26 \\
Correction & 1.60 & 1.87 \\
\bottomrule
\end{tabular}
\caption{Average additional iterations after convergence to a correct reasoning record under adaptive termination.}
\label{tab:adaptive_termination_dynamics}
\vspace{-3.0mm}
\end{table}





To assess whether adaptive termination halts once sufficient knowledge has been acquired, we measure the number of iterations executed after the reasoning process has already converged to a correct state.
We focus on \textit{Stable-Correct} and \textit{Correction} trajectories, in which a correct reasoning record is reached at some iterations and maintained thereafter.

As shown in Table~\ref{tab:adaptive_termination_dynamics}, adaptive termination typically occurs shortly after convergence on both InfoSeek and E-VQA.
Because adaptive termination requires at least one subsequent iteration to assess saturation by comparing newly generated outputs with prior states, the procedure incurs, on average, approximately two additional iterations.

\vspace{-1.0mm}
\section{Conclusion}
\vspace{-1.0mm}
In this paper, we introduced PMSR, a progressive multimodal search and reasoning framework for knowledge-intensive VQA. PMSR constructs a structured reasoning trajectory composed of compact reasoning records synthesized from diverse evidence to enhance both knowledge acquisition and synthesis. This design enables controlled, iterative refinement of retrieval and reasoning, promoting more stable trajectories that can correct early mistakes and reduce drift over successive iterations. Extensive experiments on six knowledge-intensive VQA benchmarks demonstrate consistent improvements in retrieval recall and end-to-end answer accuracy over strong baselines, highlighting the effectiveness of PMSR.

\section{Limitations}

Although the PMSR framework demonstrates significant improvements in retrieval recall and answer accuracy across several knowledge-intensive VQA benchmarks, few limitations warrant consideration for future research. First, the proposed framework relies on iterative retrieval and reasoning, which introduces additional inference overhead compared to single-pass RAG methods. Although the adaptive termination mechanism mitigates redundant iterations, the overall computational cost remains higher in cases where convergence is slow.

Second, PMSR retrieves from heterogeneous sources, but its overall performance remains sensitive to retrieval quality and query formulation. In the main experiments, we rely on standard dense retrievers rather than fully optimizing recent MLLM-based multimodal retrievers that learn fused image-text embeddings for joint retrieval. Although Appendix~\ref{sec:comparison_mllm_retrievers} provides a preliminary comparison with one such retriever, further integration with MLLM-based retrievers remains promising.

Finally, PMSR remains limited by the reasoning and grounding capabilities of the underlying MLLM. While PMSR provides progressive reasoning records and a structured, iterative knowledge acquisition process, smaller or less capable backbones may still struggle with accurate visual grounding and compositional reasoning over multiple visual-text associations, limiting how effectively retrieved relevant knowledge can be leveraged for correct predictions.

\section*{Acknowledgments}

This work was supported by Institute of Information \& communications Technology Planning \& Evaluation (IITP) grant funded by the Korea government (MSIT) ([NO.RS-2021-II211343, Artificial Intelligence Graduate School Program (Seoul National University)], [No.RS-2023-00235293, Development of autonomous driving big data processing, management, search, and sharing interface technology to provide autonomous driving data according to the purpose of usage]).

\nocite{Ando2005,andrew2007scalable,rasooli-tetrault-2015}
\bibliography{custom}


\appendix
\clearpage
\definecolor{LightGray}{gray}{0.9}

\setcounter{page}{1} 
\setcounter{figure}{0} 
\setcounter{table}{0}  
\setcounter{equation}{0}

\renewcommand{\thesection}{\Alph{section}}
\renewcommand{\thefigure}{A\arabic{figure}} 
\renewcommand{\thetable}{A\arabic{table}}   
\renewcommand{\theequation}{\thesection.\arabic{equation}} 
\section{Prompts for PMSR Framework}
\label{sec:prompts_appendix}

We present the prompt templates used in the PMSR framework. All prompts are designed to be model-agnostic. In the templates below, terms enclosed in \texttt{\{braces\}} denote dynamic content populated at runtime.

\subsection{Initial Reasoning Record Generation}
\begin{figure}[h!]
    \centering
    \begin{tcolorbox}[colback=gray!5, colframe=gray!50, title=\textbf{Initial Description Prompt}]
        \small\ttfamily
        Question: \{question\} \\
        
        \vspace{0.5em}
        \hrule
        \vspace{0.5em}
        
        \textbf{Instruction:} \\
        Concisely describe the image which is relevant to the question.
    \end{tcolorbox}
    \vspace{-3.0mm}
    \caption{The prompt used to generate the initial visual description~($d_0$) for the first query expansion.}
    \label{fig:prompt_initial}
    \vspace{-2.0mm}
\end{figure}
To bootstrap the iterative process, we first instruct the model to generate a query-focused description of the image~(Figure~\ref{fig:prompt_initial}). This serves as the initial grounding for the first retrieval step.

\subsection{Dual-scope Query Formulation}
As described in Section~\ref{sec:multi_query_formulation}, PMSR constructs the query set $\mathcal{Q}_t$ using two complementary operators: a \textit{record-level} query and a \textit{trajectory-level} query.

\paragraph{Record-level query.}
This operator conditions on the most recent reasoning record to use its most recent deductions to guide successive retrieval. It is implemented by concatenating the original question $Q$ with the latest reasoning record $r_{t-1}$, i.e., $q_{\text{record}}^{(t)} = [Q; r_{t-1}]$. No additional instruction prompt is required.

\paragraph{Trajectory-level query.}
To implement the trajectory-level query operator $\mathcal{G}_{\text{trajectory}}$, we use the structured prompt shown in Figure~\ref{fig:prompt_query_gen}. The prompt instructs the MLLM to analyze the accumulated reasoning trajectory~(provided in the \texttt{\{knowledge\}} field) together with the original question. By separating an explicit \texttt{Analysis} section from the \texttt{Output}, the model is encouraged to identify missing or underspecified information in $\langle r_0, \dots, r_{t-1}\rangle$ before generating a context-specific query for subsequent knowledge search.

\begin{figure}[h!]
    \centering
    \begin{tcolorbox}[colback=gray!5, colframe=gray!50, title=\textbf{Trajectory-level Query Operator Prompt ($\mathcal{G}_{\text{trajectory}}$)}]
        \small\ttfamily
        \textbf{Input Context:} \\
        **Query**: \{question\} \\
        **Knowledge**: \{knowledge\} \\
        
        \vspace{0.5em}
        \hrule
        \vspace{0.5em}
        
        \textbf{Instruction:} \\
        Please first analyze all the information in a section named Analysis (\#\# Analysis). \\
        Generate more accurate question based on the Knowledge to search more information helpful to addressing Query. \\
        
        Your response should be in the following format: \\
        
        \#\# Analysis \\
        Analysis query and correct knowledge to search more accurately. \\
        
        \#\# Output \\
        Question: context-specific new question
    \end{tcolorbox}
    \vspace{-3.0mm}
    \caption{The prompt used for the trajectory-level query operator $\mathcal{G}_{\text{trajectory}}$. The \texttt{\{knowledge\}} field is populated with the accumulated reasoning trajectory $\langle r_0, \dots, r_{t-1}\rangle$ to allow the model to identify gaps before generating a new search query.}
    \label{fig:prompt_query_gen}
    \vspace{-2.0mm}
\end{figure}

\subsection{Reasoning Record Generation}
At each iteration, we synthesize the retrieved evidence into a concise "Reasoning Record." This prompt is designed to retain only correct and relevant information for the next step (Figure~\ref{fig:prompt_reasoning_record}).

\begin{figure}[t!]
    \centering
    \begin{tcolorbox}[colback=gray!5, colframe=gray!50, title=\textbf{Reasoning Record Synthesis ($\mathcal{G}_{\text{reason}}$)}]
        \small\ttfamily
        \textbf{Input Context:} \\
        Question: \{question\} \\
        Knowledge: \{knowledge\} \\
        
        \vspace{0.5em}
        \hrule
        \vspace{0.5em}
        
        \textbf{Instruction:} \\
        Based on the image, description, and knowledge, summarize the correct and relevant information with respect to the image and question.
    \end{tcolorbox}
    \vspace{-3.0mm}
    \caption{The prompt used for the Reasoning Record Generation operator $\mathcal{G}_{\text{reason}}$, which synthesizes a reasoning record from newly retrieved multimodal evidence.}
    \label{fig:prompt_reasoning_record}
    \vspace{-5.0mm}
\end{figure}

\begin{figure}[h]
    \centering
    \begin{tcolorbox}[colback=gray!5, colframe=gray!50, title=\textbf{Final Answer Generation}]
        \small\ttfamily
        \textbf{Input Context:} \\
        Question: \{question\} \\
        Relevant Knowledge: \{reasoning\_records\} \\
        
        \vspace{0.5em}
        \hrule
        \vspace{0.5em}
        
        \textbf{Instruction:} \\
        Please answer the following question using the provided information and image. \\
        Based on the information, provide a detailed answer to the question.
    \end{tcolorbox}
    \vspace{-3.0mm}
    
    \caption{The final prompt used to generate the answer by synthesizing the original question, image, and the full chain of reasoning records.}
    \label{fig:prompt_final}
    \vspace{-5.0mm}
\end{figure}

\subsection{Final Answer Generation}
Once the iterations are complete (or adaptive termination is triggered), we use the reasoning trajectory to derive the final answer~(Figure~\ref{fig:prompt_final}).

\section{Prompts for Web Search}
\label{sec:web_prompts}

We employ two specialized prompts to optimize the web search process. The first ensures that search queries are concise and keyword-optimized, while the second condenses retrieved long content into text passages.

\subsection{Search Query Condensation}
When a generated query exceeds a predefined length threshold (e.g., 400 characters), we use a condensation prompt to rewrite it into a form suitable for search engines. This helps extract the main entities and essential intent from a verbose reasoning step (Figure~\ref{fig:prompt_condense}).

\begin{figure}[h!]
    \centering
    \begin{tcolorbox}[colback=gray!5, colframe=gray!50, title=\textbf{Search Query Condensation}]
        \small\ttfamily
        \textbf{Instruction:} \\
        Rewrite this query to be a concise question for search engine including main entity and essential point. \\
        
        \vspace{0.5em}
        \hrule
        \vspace{0.5em}
        
        \textbf{Input Context:} \\
        Query: \{prompt\}
    \end{tcolorbox}
    \vspace{-3.0mm}
    \caption{The prompt used to condense verbose or complex reasoning outputs into an effective keyword-based search query.}
    \label{fig:prompt_condense}
    \vspace{-5.0mm}
\end{figure}

\subsection{Web Content Summarization}
To efficiently handle the noise and length of raw web pages, we use a summarization prompt. This prompt instructs the MLLM to generate a summary relevant to the original query~(Figure~\ref{fig:prompt_summarization}).

\begin{figure}[h!]
    \centering
    \begin{tcolorbox}[colback=gray!5, colframe=gray!50, title=\textbf{Web Content Summarization}]
        \small\ttfamily
        \textbf{Input Context:} \\
        Query: \{prompt\} \\
        
        \vspace{0.5em}
        \hrule
        \vspace{0.5em}
        
        \textbf{Instruction:} \\
        Summarize the following web content, focusing on information relevant to the query. Provide a concise summary in a single paragraph: \\
        
        \vspace{0.5em}
        \hrule
        \vspace{0.5em}
        
        \textbf{Target Content:} \\
        Title: \{title\} \\
        Content: \{content\}
    \end{tcolorbox}
    \vspace{-3.0mm}
    \caption{The prompt used to summarize retrieved web pages. By explicitly conditioning the summary on the input query, the model filters irrelevant candidates and focuses on the evidence needed for the answer.}
    \label{fig:prompt_summarization}
    \vspace{-5.0mm}
\end{figure}

\section{Web-equipped Implementation}
\label{sec:web_search}

To evaluate PMSR on search-oriented benchmarks that require access to time-varying or out-of-KB information, we implement a web-equipped variant that augments PMSR with web search. This variant preserves the core PMSR pipeline (query formulation $\rightarrow$ retrieval $\rightarrow$ reasoning record generation), while adapting query construction and evidence processing to practical constraints of web search (e.g., query length limits and noisy webpage content).

\noindent \paragraph{Image search tool.}
We use the ScrapingDog API to interface with Google Lens at the initial iteration ($t{=}0$) only, in order to obtain an initial set of visually related webpages. Given the input image, the tool returns visually similar pages with metadata such as thumbnails and titles. We then fetch the corresponding page contents and generate question-conditioned summaries, which are used as the initial visual evidence to bootstrap the first reasoning record. In subsequent iterations ($t>0$), PMSR performs multimodal retrieval over the Wikipedia-based multimodal KB.

\noindent \paragraph{Text search tool.}
For textual retrieval, we employ Ollama Web Search following a \textit{search-parse-summarize} pipeline. Given a query, the system retrieves relevant URLs, parses their contents, and summarizes each page using GPT-OSS 120B (5.1B active parameters). The same model is also used to rewrite long queries into concise forms to meet search-engine constraints.

\subsection{Initial Reasoning Record Generation.}
At the initial iteration ($t=0$), we follow the protocol of MMSearch-R1 by anchoring retrieval with image-based search. Specifically, we submit the input image $I$ to Google Lens to obtain visually similar webpages, then fetch and summarize their contents to form the initial candidate set $\mathcal{D}_0$. We generate the initial reasoning record $r_0$ by applying the standard reasoning operator $\mathcal{G}_{\text{reason}}$ on $(Q, I, \mathcal{D}_0)$, thereby bootstrapping the reasoning trajectory with visually grounded evidence.

\subsection{Dual-scope Query Formulation}
In the standard PMSR framework, the record-level query is constructed by concatenating the original question with the latest reasoning record. However, web search engines impose constraints on query length and formatting, making direct concatenation impractical. To address this, we introduce a rewriting operator $\mathcal{G}_{\text{record}}$ that compresses $(Q, r_{t-1})$ into a concise search string:
\begin{equation}
q_{\text{record}}^{(t)} = \mathcal{G}_{\text{record}}(Q, r_{t-1}).
\end{equation}
The model is prompted to produce a keyword-focused query suitable for web search. For the trajectory-level query, we use the same operator $\mathcal{G}_{\text{trajectory}}$ as in Section~\ref{sec:multi_query_formulation}, which synthesizes the accumulated reasoning trajectory into a context-specific query.

\subsection{Web Search and Evidence Summarization}
Each web query is executed to retrieve a relevant list of URLs. To summarize relevant information, we apply a summarization pipeline:
\begin{enumerate}
    \item \textbf{Content extraction:} We scrape the raw HTML content of the top-$k$ retrieved webpages.
    \item \textbf{Query-conditioned summarization:} Each page is summarized to produce a relevant summary of the input query.
\end{enumerate}
The resulting text summaries and the text retrieved from image search are treated as the candidate set $\mathcal{D}_t$. We then generate the next reasoning record $r_t$ using $\mathcal{G}_{\text{reason}}(Q, I, \mathcal{D}_t)$, following the same procedure as in Section~\ref{sec:reasoning_record_generation}.

\section{Multimodal Retrieval Implementation}
\label{sec:multimodal_retrieval}
To construct the multimodal knowledge base, we process the Wikipedia corpus used in InfoSeek, which is derived from the 2022-10-01 Wikipedia dump.
For each image–text pair, we generate normalized image and text embeddings to compute a decoupled similarity score.
Specifically, image embeddings are extracted using a pretrained SigLIP2 model, while text embeddings are obtained from Wikipedia section summaries using a Qwen3-Embedding encoder.

To efficiently implement decoupled similarity retrieval, we concatenate the normalized image and text embeddings into a joint multimodal representation and index them using FAISS with an IndexFlatIP~(inner product) index.
At query time, the input image and refined text query are encoded separately using the same encoders and concatenated to form a single multimodal query vector.
A maximum inner product search~(MIPS) is then performed to retrieve the top-$k$ candidates.
With the default weight $\lambda = 0.5$, this retrieval procedure is equivalent to the decoupled similarity score, enabling efficient and scalable retrieval while preserving consistency with our scoring function.



\begin{table}[h]
\centering
\resizebox{\columnwidth}{!}
{
\begin{tabular}{c c c c c c}
\toprule
\textbf{Retriever} & \textbf{Dataset} & \textbf{\makecell{Query\\Modality}} & \textbf{R@5} & \textbf{R@10} & \textbf{R@20} \\
\midrule
\multirow{2}{*}{\makecell{OMGM}} & InfoSeek & image-to-text & 73.9 & 80.0 & 84.8 \\
\cmidrule(l){2-6}
& E-VQA & image-to-text & 41.2 & 49.8 & 58.7 \\
\midrule
\multirow{2}{*}{\makecell{EVA-CLIP-8B}} & InfoSeek & image-to-image & 67.1 & 73.0 & 77.9 \\
\cmidrule(l){2-6}
& E-VQA & image-to-image & 31.3 & 41.0 & 48.8 \\
\midrule
\multirow{4}{*}{\makecell{SigLIP2-g\\Qwen3-Embedding-0.6B}} & \multirow{2}{*}{InfoSeek} & image-to-image & 66.7 & 72.7 & 77.5 \\
& & image+text & 69.4 & 76.2 & 81.1 \\
\cmidrule(l){2-6}
& \multirow{2}{*}{E-VQA} & image-to-image & 36.2 & 41.9 & 46.4 \\
& & image+text & 43.1 & 48.7 & 54.5 \\
\bottomrule
\end{tabular}
}
\caption{
    Performance of multimodal similarity on the InfoSeek validation split and E-VQA test split. 
}
\label{tab:multimodal_similarity}
\end{table}

Table~\ref{tab:multimodal_similarity} reports retriever performance on the InfoSeek validation split and the E-VQA test split. The results show that multimodal queries that jointly incorporate image and text similarity consistently achieve higher recall than unimodal~(image-only) queries.


%
%

\section{Details of Datasets}
\label{sec:dataset}

\paragraph{InfoSeek.}

InfoSeek is a large-scale benchmark designed to evaluate visual information-seeking capabilities. It consists of automatically generated and human-annotated questions grounded in Wikidata entities, paired with corresponding images and factual answers. Question templates cover hundreds of relational types, ensuring broad coverage of entity attributes, locations, and fine-grained factual properties. Each question–image pair is retained only when supporting evidence exists in Wikipedia, resulting in a dataset well aligned with real-world encyclopedic knowledge. Following the evaluation protocol of PreFLMR, we evaluate on 5K questions from the M2KR subset of the InfoSeek validation split.

\paragraph{Encyclopedic VQA.}

Encyclopedic VQA focuses on fine-grained entity understanding across natural and landmark categories. Each entity is associated with multiple images and supported by textual evidence drawn from a large, controlled Wikipedia-derived knowledge base. The dataset includes both single-hop and multi-hop questions, enabling evaluation of visual grounding combined with factual reasoning. Consistent with prior work, we evaluate on the official E-VQA test split using only single-hop questions, which provides a clean setting for assessing retrieval quality and answer accuracy via the BEM metric.

\paragraph{FVQA.}

FVQA is a curated evaluation set of 2K questions constructed to emphasize factual reasoning grounded in visual evidence. It combines three sources: human-verified samples selected from an automatically generated FVQA pool, re-annotated examples drawn from the InfoSeek Human split, and newly collected instances by human annotators. Together, these subsets span diverse categories of factual knowledge, requiring the model to jointly interpret the image content and retrieve the appropriate supporting fact. This controlled, carefully validated setup allows precise assessment of factual multimodal reasoning. We report results on 1.8K questions from the test split.

\paragraph{OK-VQA.}

OK-VQA evaluates knowledge-intensive question answering, where answers cannot be derived from the image alone. Questions cover common-sense, cultural, geographic, and scientific knowledge, requiring external information sources to supplement visual understanding. The dataset is widely used to benchmark retrieval-augmented visual reasoning, as models must identify the relevant factual concept and connect it to the visual context in order to generate correct answers. We report results on 5K questions from the validation split, which is commonly used for benchmarking retrieval-augmented VQA systems.

\paragraph{InfoSeek Human.}

The InfoSeek Human subset, composed of 2K questions used in MMSearch-R1, was drawn from the InfoSeek Human split that demands open-domain retrieval. These samples include entity-level and relational queries where relevant evidence must be located across large textual corpora. The subset captures the retrieval-intensive aspects of InfoSeek while removing questions whose answers can be inferred solely from the image, providing a targeted test bed for evaluating search and reasoning under multimodal constraints.

\paragraph{LiveVQA.}

LiveVQA evaluates real-world information-seeking under time-sensitive news contents. The benchmark is built from contemporary articles across major global news outlets, each paired with images and automatically generated questions that range from basic visual recognition to multi-hop reasoning over the article’s text. Its emphasis on up-to-date events, diverse categories, and mixed reasoning styles makes LiveVQA an effective test of a model’s ability to retrieve current information and integrate it with visual cues. We evaluate performance on the 3,602 questions from the preview split, covering all news categories and reasoning types.

\paragraph{MMSearch.}

MMSearch contains manually curated examples spanning a wide range of real-world domains, divided into knowledge-oriented and news-oriented queries. The benchmark includes a subset of visual questions that require models to perform multimodal retrieval over both general knowledge and rare, specialized facts. Many questions are chosen specifically because leading LLMs struggle to answer them without external search. This makes MMSearch particularly suitable for evaluating agentic or iterative retrieval systems designed for complex information-seeking tasks. For evaluation, we use the visual subset of 171 questions, which isolates multimodal information-seeking scenarios requiring retrieval beyond image content.

\section{Prompts for LLM-as-a-Judge}
\label{sec:prompt_for_judge}

\begin{figure}[t!] 
    \centering
    \begin{tcolorbox}[colback=gray!5, colframe=gray!50, title=\textbf{Prompt Template for LLM-as-a-Judge}]
        \small\ttfamily 
        
        \textbf{Input Format:} \\
        Question: \{question\} \\
        Ground Truth Answers: \{gold\_answer\} \\
        Model Response: \{model\_response\}
        
        \vspace{0.5em}
        \hrule
        \vspace{0.5em}
        
        \textbf{Evaluation Instructions:} \\
        You are an AI assistant tasked with evaluating the correctness of model responses given the Question and Ground Truth answer. Your judgment should follow these principles:
        \begin{enumerate}[leftmargin=*, noitemsep, topsep=0pt]
            \item Consider the question, and ground truth answer holistically before evaluating the model's response.
            \item Your decision should be strictly \textbf{Yes or No}, based on whether the model's response is factually accurate and aligns with the ground truth answer.
            \item If the model response is a more specific form that includes the ground truth answer, it is correct.
            \item If the model response includes all key information but adds minor details, it is correct as long as the extra details are factually correct.
            \item If the model response contradicts, modifies, or omits critical parts of the answer, it is incorrect.
            \item For numerical values, ensure correctness even when presented in different units.
            \item For names, check for first and last name correctness. If the middle name is extra but correct, consider it correct.
            \item For yes/no questions, the response must exactly match "Yes" or "No" to be correct.
        \end{enumerate}
        Evaluate whether the Model Response is correct based on the Question and Ground Truth Answer. Follow the predefined judgment rules and provide a clear Yes/No answer along with a justification.
        
        \vspace{0.5em}
        \hrule
        \vspace{0.5em}

        \textbf{Output Format:} \\
        <reason>Detailed reasoning following the evaluation principles.</reason> \\
        <judge>Yes/No</judge>
        
    \end{tcolorbox}
    \vspace{-3.0mm}
    \caption{\textbf{Prompt template for LLM-as-a-Judge evaluation.} We employ this structured prompt to enforce strict factual consistency while allowing minor semantic variations. The placeholders \texttt{\{question\}}, \texttt{\{gold\_answer\}}, and \texttt{\{model\_response\}} are populated dynamically for each sample.}
    \label{fig:judge_prompt}
    \vspace{-5.0mm}
\end{figure}

For OK-VQA, FVQA, InfoSeek Human, MMSearch, and LiveVQA, we follow the LLM-as-a-Judge evaluation framework introduced in MMSearch-R1~\cite{wu2025mmsearch}.  Given an input question, a ground-truth answer, and a model prediction, a judging LLM evaluates whether the prediction is correct, producing both a binary decision and a brief justification. The evaluation focuses on semantic equivalence rather than exact string matching, while enforcing strict correctness for core factual content, names, and numerical values. Concretely, we use the prompt template shown in Figure~\ref{fig:judge_prompt}, where \texttt{\{question\}}, \texttt{\{gold\_answer\}}, and \texttt{\{model\_response\}} are filled with the corresponding values for each sample.

\section{Per-iteration Recall Analysis}
\label{sec:hetero_kb_iteration}
\begin{table}[h]
\centering
\small
\setlength{\tabcolsep}{5pt}
\resizebox{\columnwidth}{!}
{
\begin{tabular}{l l c c c c}
\toprule
\multirow{2}{*}{\textbf{Dataset}} & \multirow{2}{*}{\textbf{Knowledge Source}} & \multicolumn{4}{c}{\textbf{Recall}} \\
\cmidrule(lr){3-6}
 & & \textbf{Iter. 1} & \textbf{Iter. 2} & \textbf{Iter. 3} & \textbf{Iter. 4} \\
\midrule
\multirow{3}{*}{\textbf{InfoSeek}} 
 & Textual KB & 62.9 & 66.3 & 68.0 & 69.2 \\
 & Multimodal KB & 83.3 & 84.7 & 85.1 & 85.5 \\
 & Heterogeneous KB & 86.0 & 87.5 & 88.1 & 88.5 \\
\midrule
\multirow{3}{*}{\textbf{E-VQA}} 
& Textual KB & 41.4 & 45.5 & 47.5 & 49.2 \\
& Multimodal KB & 54.8 & 56.2 & 56.8 & 57.5 \\
& Heterogeneous KB & 63.8 & 65.5 & 66.6 & 67.5 \\
\bottomrule
\end{tabular}
}
\caption{Per-iteration recall of different knowledge sources in PMSR using Qwen3-VL-8B. Heterogeneous KB combines textual KB~(R@20) and multimodal KB~(R@10) at each iteration. }
\label{tab:hetero_kb_iter}
\end{table}

Table~\ref{tab:hetero_kb_iter} reports a per-iteration recall for the textual KB, multimodal KB, and their heterogeneous combination, offering a fair comparison of how each source contributes during iterative retrieval. Across both InfoSeek and E-VQA, each knowledge source shows incremental gains over iterations, while the heterogeneous setting consistently achieves the highest recall at every step. These results indicate that PMSR’s iterative refinement leverages diverse associations from both sources, and that its improvements arise from their complementary signals.

\section{Accuracy on OK-VQA}
\label{sec:OKVQA_Accuracy}
\begin{table}[t]
\centering
\resizebox{1.0\columnwidth}{!}{
\begin{tabular}{l c c}
\toprule
\textbf{Method} & \textbf{Knowledge Source} & \textbf{OK-VQA} \\
\midrule
MMSearch-R1-7B~\cite{wu2025mmsearch} & \makecell{Google Search\\Google Lens} & 59.9 \\
DeepMMSearch-R1-7B~\cite{narayan2025deepmmsearch} & \makecell{Google Search\\Google Lens} & 67.8 \\
\midrule
PMSR (Qwen3-VL-8B) & Wikipedia & 66.0 \\
\bottomrule
\end{tabular}
}
\caption{Accuracy comparison on the OK-VQA benchmark, comparing our Wikipedia-based approach with the web-equipped baselines from~\cite{narayan2025deepmmsearch}. }
\label{tab:okvqa_comparison}
\end{table}
To assess synthesis quality beyond retrieval recall, we further evaluate end-to-end accuracy on OK-VQA using LLM-as-a-Judge protocol. Given the open-ended nature of this benchmark, we compare PMSR against agentic systems that utilize live web search tools. As shown in Table~\ref{tab:okvqa_comparison}, PMSR achieves competitive performance using only the Wikipedia knowledge source, recording 66.0\% accuracy with the Qwen3-VL-8B backbone. 

\section{Record-level Query Analysis}
\begin{table}[t]
\centering
\small
\begin{tabular}{lccc}
\toprule
\textbf{Text Signal} & \textbf{R@5} & \textbf{R@10} & \textbf{R@20} \\
\midrule
Question only & 43.1 & 48.7 & 54.5 \\
Question + reason. record & 46.6 & 52.1 & 56.2 \\
\bottomrule
\end{tabular}
\caption{Effect of incorporating the latest reasoning record into the record-level query of PMSR for multimodal retrieval on E-VQA, using only the multimodal KB with the default weight $\lambda{=}0.5$. }
\label{tab:rr_vs_question}
\end{table}
\label{sec:record_ablation}

Table~\ref{tab:rr_vs_question} compares retrieval over the multimodal KB using the question alone versus the record-level query that appends the latest reasoning record. 
Incorporating the reasoning record consistently improves Recall@5/10/20 (+3.5/+3.4/+1.7 points).
These results indicate that the reasoning record provides an additional reasoning-guided signal that helps retrieve more relevant knowledge.

\section{Scaling Multimodal Retrievers}
\label{sec:scaling_retrievers}
\begin{table}[t]
\scriptsize
\centering
\begin{tabular}{l|c|cc}
\toprule
\textbf{Model} & \textbf{\makecell{Retriever Size}} & \textbf{BEM} & \textbf{Recall} \\ \midrule
\multirow{2}{*}{Qwen3-VL-4B} & Small & 39.8 & 62.5 \\
& Large & 40.9 & 68.1 \\ \midrule
\multirow{2}{*}{Qwen3-VL-8B} & Small & 45.5 & 63.4 \\
& Large & 47.1 & 67.5 \\ \bottomrule
\end{tabular}
\caption{Impact of retriever scaling on E-VQA, comparing a small retriever (SigLIP2-SO400m + ModernBert-GTE) and a large retriever (SigLIP2-g + Qwen3-Emb). Results are reported under the fixed 5-iteration setting.}
\label{tab:retriever_scaling}
\end{table}

We investigate the impact of retriever capacity on overall performance. To this end, we compare a small retriever (SigLIP2-So400m + ModernBERT-GTE) with a large retriever (SigLIP2-giant + Qwen3-Embedding-0.6B). As summarized in Table~\ref{tab:retriever_scaling}, this scaling yields consistent gains in recall on the E-VQA benchmark. Crucially, these improvements in retrieval show increases in BEM accuracy across both model sizes.

\section{Top-k Sensitivity Analysis}
\begin{table}[h]
\centering
\scriptsize
\resizebox{\columnwidth}{!}
{
\begin{tabular}{l c c c}
\toprule
\textbf{Model} & \textbf{Metric} & \textbf{$k=10$} & \textbf{$k=20$} \\
\midrule
\multirow{2}{*}{Qwen3-VL-4B} 
& BEM   & 40.9    & 42.4 \\
& Recall  & 64.3    & 72.7 \\
\cmidrule(lr){1-4}
\multirow{2}{*}{Qwen3-VL-8B} 
& BEM   & 46.4    & 46.3 \\
& Recall  & 67.3    & 72.8 \\
\bottomrule
\end{tabular}
}
\caption{Top-$k$ sensitivity analysis of PMSR on the E-VQA benchmark, showing the impact of retrieval budget on different Qwen3-VL models. $k$ denotes the number of retrieved image-text pairs.}
\label{tab:top_k_sensitivity}
\end{table}

Table \ref{tab:top_k_sensitivity} analyzes the effect of increasing the retrieval budget on performance on E-VQA. Expanding the number of retrieved image–text pairs from $k=10$ to $k=20$ consistently improves retrieval recall for both Qwen3-VL-4B and Qwen3-VL-8B, indicating increased evidence coverage. However, answer accuracy shows marginal gains for the 4B model and remains largely unchanged for the 8B model.

\section{Sensitivity of the Multimodal Similarity Weight}
\label{sec:lambda_ablation}
\begin{table}[h!]
\centering
\small
\begin{tabular}{c ccc}
\toprule
\textbf{$\lambda$} & \textbf{R@5} & \textbf{R@10} & \textbf{R@20} \\
\midrule
0.3 & 48.2 & 52.8 & 57.0 \\
0.4 & 47.9 & 52.7 & 57.0 \\
0.5 & 46.6 & 52.1 & 56.2 \\
0.6 & 44.6 & 49.7 & 54.6 \\
0.7 & 42.5 & 47.4 & 52.4 \\
\bottomrule
\end{tabular}
\caption{Ablation of the multimodal similarity weight λ for balancing text and image similarity using latest reasoning record on E-VQA.}
\label{tab:lambda_rr_ablation}
\end{table}





Table~\ref{tab:lambda_rr_ablation} reports Recall@5/10/20 for $\lambda \in \{0.3, 0.4, 0.5, 0.6, 0.7\}$. Performance peaks at smaller $\lambda$ (0.3-0.4) and gradually decreases as $\lambda$ increases, suggesting that maintaining visual relevance is important. At the same time, recall remains competitive across a broad range of $\lambda$, indicating that the record-level query provides a useful text signal that complements visual matching.

\section{Sensitivity to Contextual Noise}
\label{sec:contextual_noise}
\begin{table}[t]
\centering
\small
\setlength{\tabcolsep}{8pt}
\resizebox{\columnwidth}{!}
{
\begin{tabular}{lc}
\toprule
\textbf{Context Configuration} & \textbf{Accuracy} \\
\midrule
Oracle Section Text & 86.6\% \\
Oracle + 10 Retrieved Pairs & 78.7\% \\
Oracle + 10 Retrieved Pairs (w/ Web Search) & 72.5\% \\
Oracle + 20 Retrieved Pairs (w/ Web Search) & 64.3\% \\
\bottomrule
\end{tabular}
}
\caption{Sensitivity analysis on contextual noise in E-VQA using Qwen3-VL-8B. All configurations include the ground-truth oracle section text. Retrieved pairs denote image-text pairs from the multimodal KB, whereas web-search rows additionally incorporate web-retrieved textual evidence. The results show that additional retrieved context can introduce distractors and degrade accuracy even when oracle evidence is available.}
\label{tab:distraction_sensitivity}
\end{table}

To further examine the gap observed in Section~\ref{subsec:reasoning_trajectory}, we analyze the model’s sensitivity to contextual noise introduced by retrieved \emph{distractors}. In particular, we test whether adding extra retrieved context can degrade answer prediction even when the oracle evidence is already present in text passages.

We conduct a controlled sensitivity analysis on the E-VQA subset, restricting the evaluation to samples for which oracle textual evidence is exactly available. This setup allows us to isolate the impact of distracting context while holding the presence of correct supporting evidence constant. As summarized in Table~\ref{tab:distraction_sensitivity}, conditioning the model solely on the oracle text yields an accuracy of 86.6\%. However, augmenting this context with retrieved image–text pairs reduces accuracy to 78.7\%, suggesting that visually similar but semantically irrelevant images can interfere with correct entity grounding. When additional textual context retrieved from Google Search is further incorporated, accuracy decreases to 72.5\%. Overall, the results highlight that contextual distractors can substantially impair evidence utilization even when correct supporting text is available.

\section{Trajectory-Type Comparison on E-VQA}
\label{sec:traj_comp_webwatcher}

\begin{table}[t]
\centering
\small
\setlength{\tabcolsep}{5.5pt}
\resizebox{\columnwidth}{!}
{
\begin{tabular}{lcccc}
\toprule
Method & Stable-Correct & Correction & Conflicts & Persistent-Fail \\
\midrule
PMSR (Ours)  & 47.7 & 30.1 & 17.3 & 4.9 \\
WebWatcher   & 12.1 & 31.2 & 4.9 & 51.8 \\
\bottomrule
\end{tabular}
}
\caption{Trajectory-type distribution on E-VQA test split. Trajectory types are defined by per-iteration correctness of records (BEM): Stable-Correct (correct at all iterations), Persistent-Fail (incorrect at all iterations), Correction (recovers from incorrect reasoning and is correct at the final iteration), Conflicts (correct in some iterations but incorrect at the final iteration).}
\label{tab:traj_comp_evqa_webwatcher}
\end{table}

To better understand the behavioral differences between progressive record-based updating and global-trajectory-only updating under the same KB and retriever, we compare the distributions of reasoning trajectory types on E-VQA between PMSR and WebWatcher.



WebWatcher exhibits a substantial \textit{Correction} rate~(31.2\%), indicating that it can revise its trajectory and recover from some initially incorrect states. However, it also shows a high \textit{Persistent-Fail} rate~(51.8\%), meaning that many examples remain incorrect across all iterations. This observation indicates that early failure steps can persist across iterations and continue to influence subsequent actions and reasoning, which may make some initial failures difficult to overcome. Supporting this, \textit{Persistent-Fail} trajectories in WebWatcher take additional iterations on average~(2.85 extra iterations, ranging from 1 to 9) without improving final correctness, suggesting that more steps do not necessarily enable recovery in these cases. In contrast, PMSR exhibits a much lower \textit{Persistent-Fail} rate~(4.9\%) and higher \textit{Stable-Correct} rate~(47.7\%), while maintaining a comparable \textit{Correction} rate~(30.1\%).

\section{Qualitative Examples}
\label{sec:qualitative_examples}

This section presents qualitative examples illustrating how PMSR progressively formulates dual-scope queries and constructs structured reasoning records over iterations. Examples are drawn from InfoSeek, FVQA, and E-VQA, using PMSR instantiated with Qwen3-VL-8B. Each figure corresponds to one case and visualizes the iterative trajectory~(reasoning records, dual-scope queries, and prediction).

To improve readability, we condense each reasoning trajectory in the figures by retaining at most two representative updates~(i.e., up to $t \le 2$) and omitting minor intermediate details. Specifically, we preserve the key transitions that drive progressive search and reasoning: (i) the initial record that bootstraps the trajectory, (ii) an intermediate update where dual-scope queries retrieve new evidence that revises or sharpens reasoning, and (iii) the final update that resolves the question. For each retained step, we report essential points of the reasoning record~(grounded entities, newly retrieved facts, and the resulting inference), while omitting auxiliary text such as partial evidence lists, redundant descriptions, and formatting artifacts. This condensed presentation highlights how PMSR progressively refines its retrieval and reasoning across iterations.

\begin{figure*}[h]
    \centering
    \includegraphics[width=0.92\textwidth]{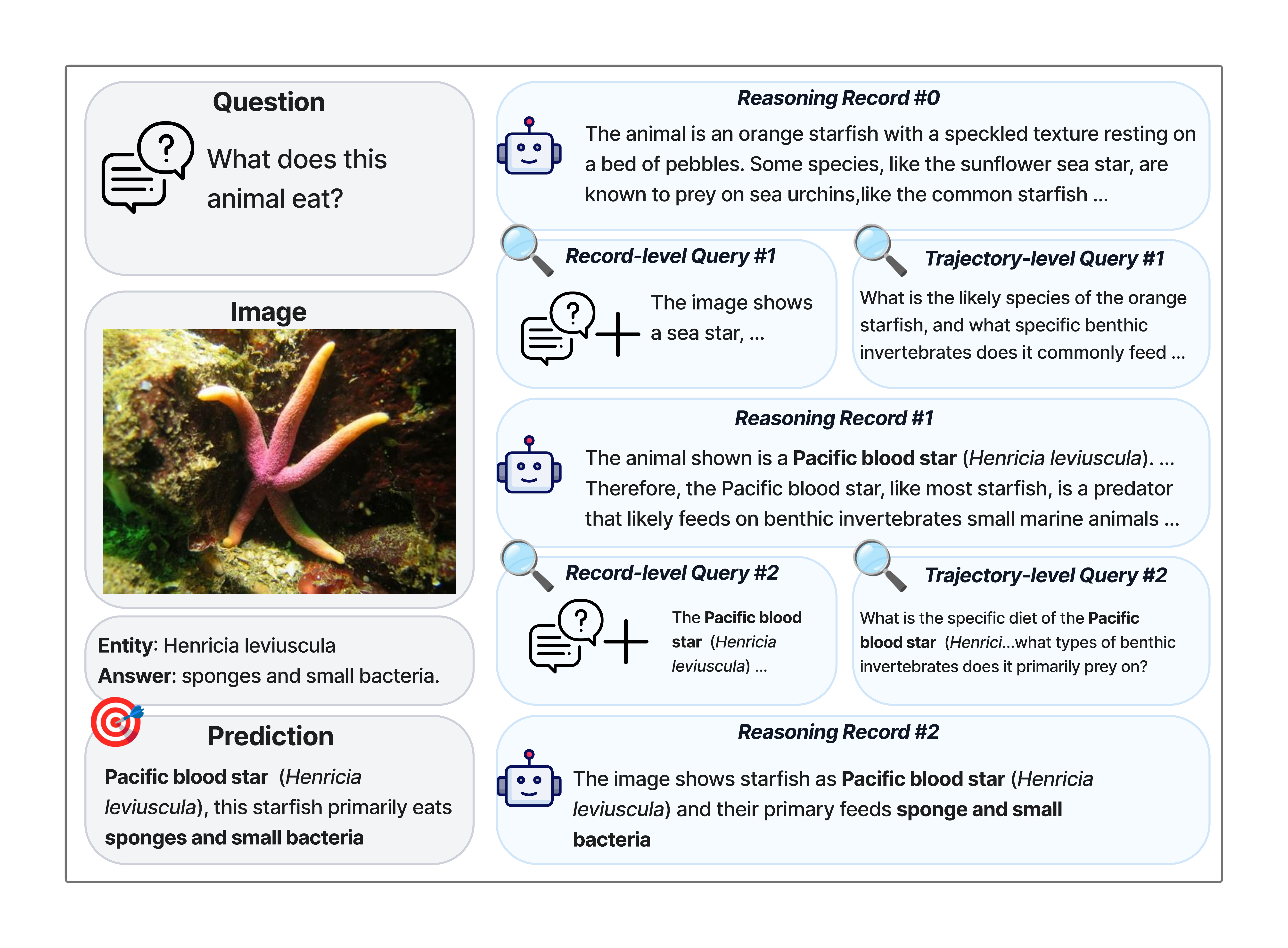}
    \caption{\textbf{E-VQA case: diet of a sea star.}
    PMSR progressively refines visual grounding and retrieves entity-specific evidence via dual-scope queries, enabling the reasoning records to converge to the correct diet.}
    \label{fig:qual_ex1}
    \vspace{-0.8em}
\end{figure*}

\paragraph{Case 1~(E-VQA: diet of a sea star).}
As shown in Figure~\ref{fig:qual_ex1}, the initial reasoning record $r_0$ relies on generic sea-star knowledge and contains only loosely related evidence, which is insufficient to answer the question. 
In subsequent iterations, PMSR decomposes retrieval into dual scopes: the record-level query targets the most recent uncertainty by refining species-level grounding, while the trajectory-level query preserves the overall intent of retrieving diet knowledge for the grounded entity. 
This reasoning-guided retrieval surfaces species-specific passages for \emph{Pacific blood star}~(\emph{Henricia leviuscula}), enabling PMSR to update the record with precise dietary information and converge on the correct answer, \emph{sponges and small bacteria}.%

\begin{figure*}[h]
    \centering
    \includegraphics[width=0.92\textwidth]{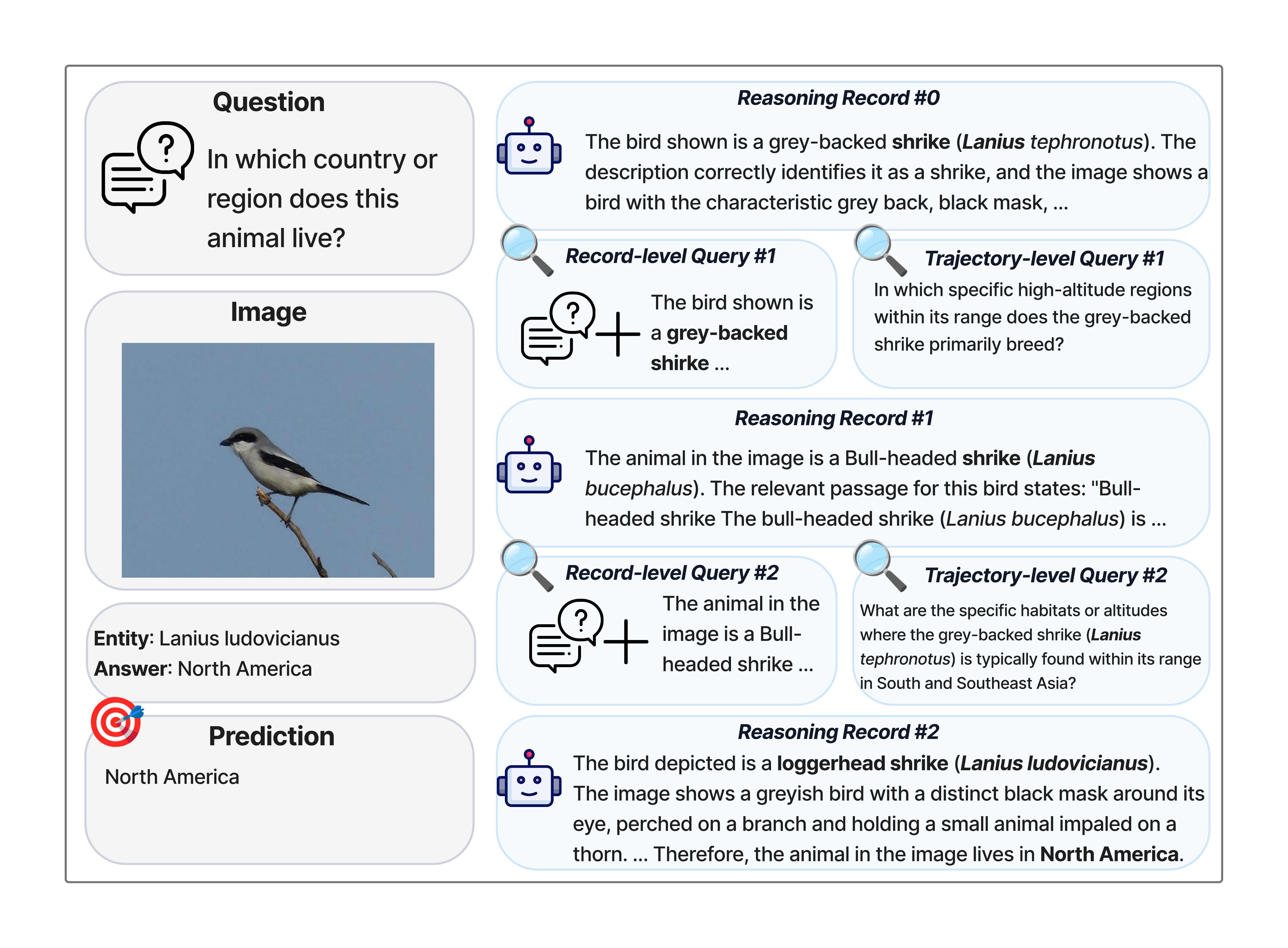}
    \caption{\textbf{E-VQA case: geographic region of a shrike.}
    Dual-scope queries mitigate early mis-grounding among visually similar species and retrieve the knowledge of the grounded entity.}
    \label{fig:qual_ex2}
    \vspace{-0.8em}
\end{figure*}

\paragraph{Case 2~(E-VQA: geographic region of a shrike).}
This example is challenging due to visually similar shrike species, which can induce errors in early grounding and retrieval (Figure~\ref{fig:qual_ex2}).
Across iterations, PMSR retrieves knowledge of specific species that better match the visual cues, enabling later records to recover from early confusion and finalize the correct region~(\emph{North America}).

\begin{figure*}[h]
    \centering
    \includegraphics[width=0.92\textwidth]{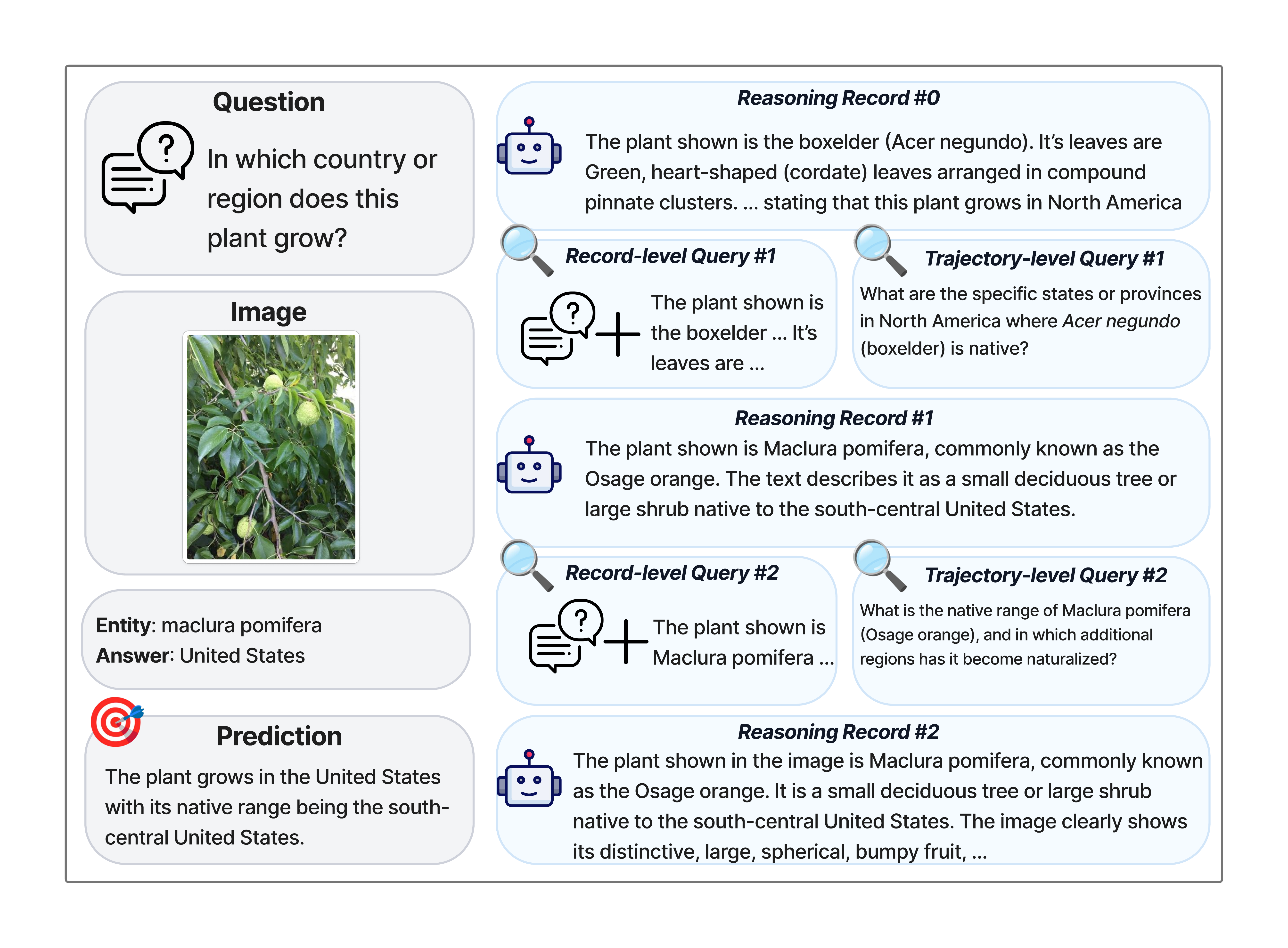}
    \caption{\textbf{E-VQA case: native range of a plant.}
    PMSR progressively aligns retrieved encyclopedic evidence with discriminative visual attributes to resolve the plant identity and its native distribution.}
    \label{fig:qual_ex3}
    \vspace{-0.8em}
\end{figure*}

\paragraph{Case 3~(E-VQA: native range of a plant).}
The initial record exhibits an early grounding failure, identifying an incorrect visual entity (e.g., \emph{boxelder}), which leads to only coarse and partially mismatched regional knowledge~(Figure~\ref{fig:qual_ex3}). Subsequent dual-scope queries improve retrieval toward discriminative visual attributes~(e.g., the distinctive large, spherical fruit) while maintaining the trajectory’s objective of resolving the plant’s native range.
Across iterations, PMSR identifies the plant as \emph{Maclura pomifera}~(Osage orange) and describes its native distribution, enabling later records to correct the initial grounding and retrieve corresponding knowledge. The final trajectory converges to the correct answer~(\emph{United States}), illustrating how PMSR recovers from early grounding errors through progressive retrieval and record updates.

\begin{figure*}[h]
    \centering
    \includegraphics[width=0.87\textwidth]{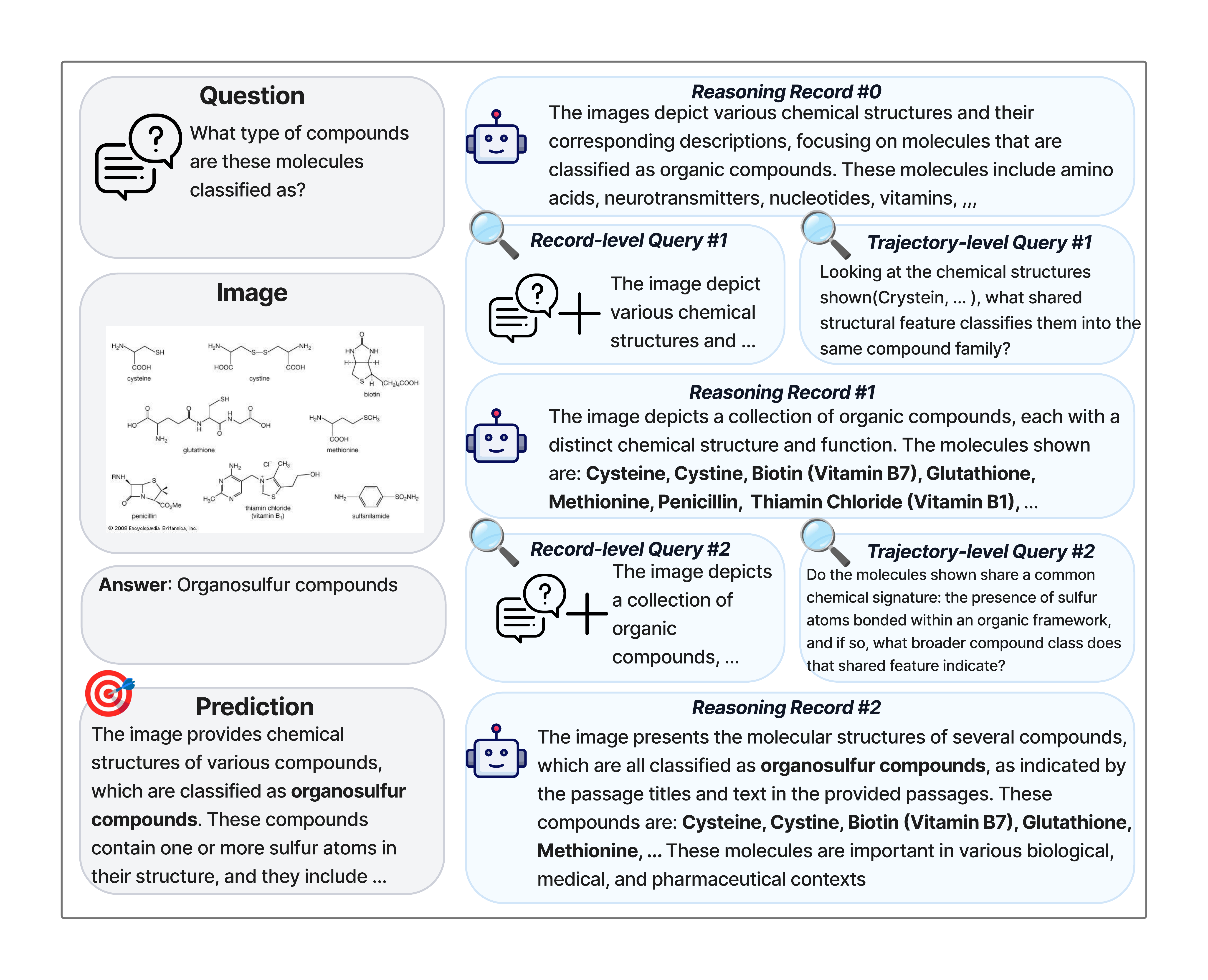}
    \caption{
    \textbf{FVQA case: chemical class of molecules.}
    PMSR narrows the classification from an overly broad concept to the targeted class by retrieving knowledge about shared structural motifs and updating the reasoning record accordingly.}
    \label{fig:qual_ex4}
    \vspace{-0.8em}
\end{figure*}

\paragraph{Case 4~(FVQA: chemical class of molecules).}
The initial reasoning record provides a broad classification that is correct but underspecified for the question (Figure~\ref{fig:qual_ex4}).
PMSR then uses dual-scope querying: the record-level query seeks discriminative evidence~(i.e., the shared structural signature), while the trajectory-level query focuses on the depicted molecules that share a common motif.
Across iterations, PMSR retrieves evidence highlighting that the depicted molecules contain sulfur atoms within an organic framework, enabling the synthesized record to resolve the intended class~(\emph{organosulfur compounds}). This example illustrates how PMSR refines from generic to specific knowledge through iterative retrieval and reasoning-record updates.

\begin{figure*}[h]
    \centering
    \includegraphics[width=0.87\textwidth]{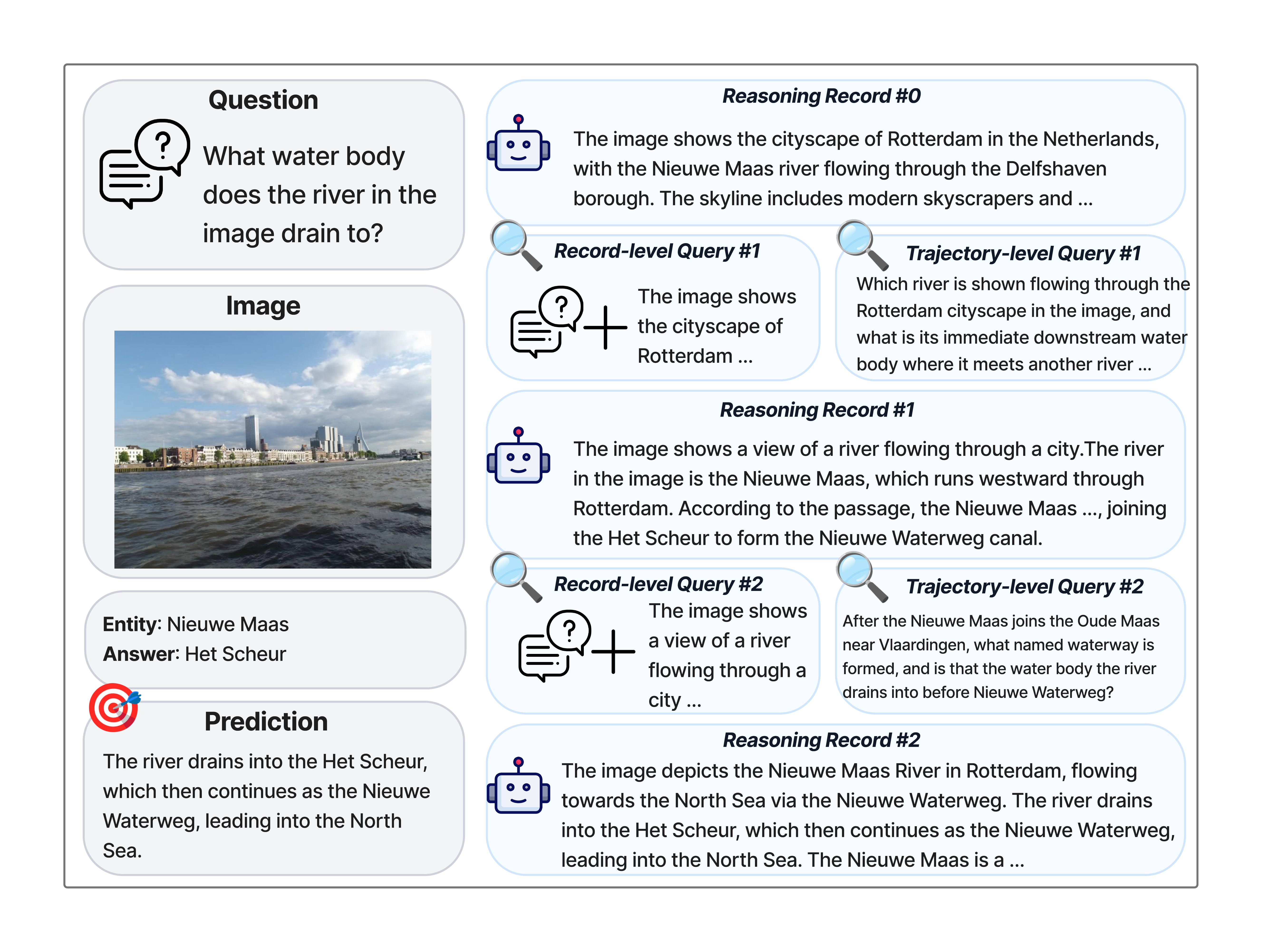}
    \caption{\textbf{InfoSeek case: downstream water body of an urban river.}
    PMSR refines visual grounding from coarse geographic grounding (Rotterdam cityscape) to a fine-grained prediction of the river’s immediate downstream water body.}
    \label{fig:qual_ex5}
    \vspace{-0.8em}
\end{figure*}

\paragraph{Case 5~(InfoSeek: downstream water body of an urban river).}
In Figure~\ref{fig:qual_ex5}, the initial record grounds the scene as the Rotterdam cityscape and identifies the river as the Nieuwe Maas, but the question requires the \emph{immediate} water body it drains into rather than the eventual outlet. In subsequent iterations, PMSR’s record-level query focuses on confirming the river identity and its downstream connection, while the trajectory-level query targets the broader river-network relation. Across iterations, PMSR retrieves knowledge corresponding to the Nieuwe Maas river system, indicating that it joins the Oude Maas near Vlaardingen and drains into the Het Scheur, which then continues as the Nieuwe Waterweg toward the North Sea. Synthesizing this knowledge, the latest reasoning record resolves the intended answer as \emph{Het Scheur}.

\clearpage

\clearpage

\section{Runtime Latency Analysis}
\label{sec:additional_analysis}

To quantify the overhead of progressive search and reasoning, we report wall-clock latency under the same inference setting used in our main experiments. Unless otherwise stated, all times are reported in seconds. We first measure a na\"ive sequential implementation on a commodity server with a single RTX~3090, and then evaluate two practical optimizations: concurrent retrieval across heterogeneous KBs and faster VLM decoding on a higher-capability GPU.

\subsection{Runtime Latency Breakdown}
\label{sec:latency_breakdown}

\begin{table}[t]
\centering
\small
\resizebox{\columnwidth}{!}
{
\begin{tabular}{llc}
\toprule
Operation Type & Component & Time (s) \\
\midrule
Reasoning & Dual-Scope Query Formulation & 3.34 \\
Retrieval & Two Text Index Retrievals & 5.51 \\
Retrieval & Two Multimodal Index Retrievals & 2.83 \\
Reasoning & Reasoning Record Generation & 5.52 \\
\bottomrule
\end{tabular}
}
\caption{Component-wise latency breakdown of PMSR for one reasoning iteration under a vanilla implementation.}
\label{tab:latency_breakdown}
\end{table}
\begin{table}[t]
\centering
\small
\resizebox{\columnwidth}{!}
{
\begin{tabular}{lcc}
\toprule
Setting & RAG (1-step) & PMSR (1-iter.) \\
\midrule
Baseline & 3.91 & 17.20 \\
+ Concurrent retrieval & 3.91 & 11.61 \\
+ H100 GPU decoding & 2.14 & 4.33 \\
\bottomrule
\end{tabular}
}
\caption{Runtime comparison between single-step RAG and PMSR. All times are reported in seconds.}
\label{tab:runtime_comparison}
\end{table}

Under the na\"ive sequential setting, the per-iteration runtime is 17.20 seconds. Table~\ref{tab:latency_breakdown} provides a component-wise latency analysis of PMSR for one reasoning iteration, showing that retrieval and VLM decoding account for most of the latency.

\subsection{Mitigating Runtime Latency}
\label{sec:mitigate_latency}

Table~\ref{tab:runtime_comparison} summarizes the effect of two practical runtime optimizations for PMSR. First, retrieval calls can be parallelized across both dual-scope query and heterogeneous KBs. This reduces retrieval wall-clock time from 8.34 seconds in the sequential implementation to approximately 2.75 seconds, lowering the per-iteration latency from 17.20 to 11.61 seconds without changing the algorithm itself.

Second, VLM decoding latency can be accelerated by using a higher-capability GPU (Nvidia H100). Using a higher-capability GPU for VLM decoding reduces the latency of Qwen3-VL-8B from 8.86 seconds to 1.58 seconds per sample. Overall, these practical optimizations reduce PMSR latency to 4.33 seconds per iteration, substantially narrowing the gap to single-step RAG. With adaptive termination, PMSR executes 3.5 iterations on average, corresponding to an end-to-end runtime of approximately 15.15 seconds per sample.

\section{Comparison of Record-level Query Design}
\label{sec:dual_scope_query_design}

\begin{table}[t]
\centering
\small
\resizebox{\columnwidth}{!}
{
\begin{tabular}{llcc}
\toprule
\makecell{Record-level\\query operator} & Dataset & Acc. ($\uparrow$) & Recall ($\uparrow$) \\
\midrule
Concatenation & InfoSeek & 50.9 & 87.6 \\
Entity extraction & InfoSeek & 52.9 & 86.6 \\
Concatenation & E-VQA & 43.5 & 66.7 \\
Entity extraction & E-VQA & 41.0 & 64.9 \\
\bottomrule
\end{tabular}
}
\caption{Comparison of record-level query operators. Results are reported on subsets of the InfoSeek validation set and the E-VQA test split for efficiency.}
\label{tab:dual_scope_record_query}
\end{table}

We also analyze the design of the record-level query transformation. In PMSR, the record-level query is formed by concatenating the original question with the latest reasoning record. To evaluate whether a complex transformation is beneficial, we additionally compare concatenation with an entity-extraction-based operator on InfoSeek and E-VQA. As shown in Table A12, the two operators yield comparable results on InfoSeek, while concatenation performs better on E-VQA. These results suggest that a more complex query transformation is not consistently beneficial in our setting. One possible explanation is that the latest reasoning record already encodes the key deductions produced by compositional reasoning; transforming it into an alternative query form may discard useful context or introduce noise, rather than reliably improving retrieval.

\section{Analysis of Compositional Reasoning in Reasoning Record Generation}
\label{sec:compositional_reasoning_record_generation}

\begin{table}[t]
\centering
\small
\resizebox{\columnwidth}{!}
{
\begin{tabular}{lcccc}
\toprule
Dataset & Filter & Summarize & Compose & \makecell{Resolve\\Conflicts} \\
\midrule
InfoSeek & 19.0 & 92.0 & 48.0 & 62.0 \\
E-VQA & 23.5 & 86.0 & 50.5 & 70.5 \\
\bottomrule
\end{tabular}
}
\caption{LLM-as-a-judge analysis of behaviors exhibited by reasoning records generated by $G_{\text{reason}}$. Each trajectory can be assigned multiple labels, since its reasoning records can jointly exhibit filtering, summarization, composition, and resolving conflicts across iterations.}
\label{tab:greason_behavior}
\end{table}

To assess whether the $G_{\text{reason}}$ operator performs compositional reasoning, we conduct an LLM-as-a-judge analysis over reasoning records generated by PMSR trajectories. For each sample, the judge assigns one or more behavior labels from \emph{Filter}, \emph{Summarize}, \emph{Compose}, and \emph{Resolve Conflicts}. Here, \emph{Compose} denotes combining multiple pieces of evidence into a conclusion not explicitly stated in any single retrieved item, while \emph{Resolve Conflicts} denotes identifying or reconciling inconsistent evidence across retrieved candidates.

As shown in Table~\ref{tab:greason_behavior}, \emph{Summarize} is common, as the reasoning record consolidates retrieved knowledge into a compact state. At the same time, we observe substantial rates of \emph{Compose} and \emph{Resolve Conflicts}: 48.0\% and 62.0\% on InfoSeek, and 50.5\% and 70.5\% on E-VQA, respectively. These results indicate that $G_{\text{reason}}$ performs compositional reasoning to synthesize information beyond summarization. 

\section{Comparison with MLLM-based Retrievers}
\label{sec:comparison_mllm_retrievers}

\begin{table}[t]
\centering
\small
\resizebox{\columnwidth}{!}
{
\begin{tabular}{lccc}
\toprule
Retriever & Dataset & Acc. ($\uparrow$) & Recall ($\uparrow$) \\
\midrule
Default retriever (ours) & InfoSeek & 50.9 & 87.6 \\
Qwen3-VL-Embedding-2B & InfoSeek & 49.1 & 82.0 \\
Default retriever (ours) & E-VQA & 43.5 & 66.7 \\
Qwen3-VL-Embedding-2B & E-VQA & 43.3 & 60.2 \\
\bottomrule
\end{tabular}
}
\caption{Comparison between the default retriever used in PMSR and an MLLM-based retriever. Results are reported on subsets of the InfoSeek validation set and the E-VQA test split for efficiency.}
\label{tab:mllm_retriever_comparison}
\end{table}

Our main experiments use standard dense retrievers for both textual and multimodal retrieval. We adopt this setting because PMSR requires pairwise retrieval over multimodal KBs, which many recent MLLM-based retrievers do not explicitly target. However, some recent MLLM-based retrievers support the pairwise retrieval capability required by PMSR. To examine whether such retrievers provide complementary gains in our framework, we additionally integrate Qwen3-VL-Embedding-2B~\cite{li2026qwen3} into PMSR and evaluate it on InfoSeek and E-VQA.

As shown in Table~\ref{tab:mllm_retriever_comparison}, the MLLM-based retriever does not consistently improve either retrieval recall or end-to-end answer accuracy over the default retriever. On InfoSeek, Qwen3-VL-Embedding-2B achieves 49.1 accuracy and 82.0 recall, compared to 50.9 and 87.6 for the default retriever; on E-VQA, it achieves 43.3 accuracy and 60.2 recall, compared to 43.5 and 66.7. These results show that, under our current setup, our retriever configuration yields better results than the similarly sized MLLM-based retriever.

\section{Generalizability Beyond Knowledge-Intensive VQA}
\label{sec:generalizability_beyond_vqa}

\begin{table}[t]
\centering
\small
\begin{tabular}{lc}
\toprule
Method & Acc. ($\uparrow$) \\
\midrule
CoT & 31.1 \\
RAG & 46.6 \\
IRCoT~\cite{trivedi2023interleaving} & 50.9 \\
Search-R1-7B~\cite{jin2025search} & 58.6 \\
s3~\cite{jiang2025s3} & 59.0 \\
PMSR (ours) & 59.8 \\
\bottomrule
\end{tabular}
\caption{Cross-domain evaluation on the HotpotQA dev set under an LLM-based evaluation protocol, following the LLM-as-a-judge protocol used in s3.}
\label{tab:hotpotqa_generalization}
\end{table}

To examine the cross-domain applicability of PMSR beyond multimodal VQA, we additionally evaluate PMSR on text-only knowledge-intensive question answering. Specifically, we use the HotpotQA~\cite{yang2018hotpotqa} dev split as a representative benchmark and adapt PMSR to the text-only setting while preserving its core design: progressive retrieval, record-level and trajectory-level query formulation, and structured reasoning-state updates over a textual KB only.

As shown in Table~\ref{tab:hotpotqa_generalization}, PMSR achieves competitive performance on HotpotQA and outperforms several iterative retrieval baselines. These results suggest that the proposed framework is applicable beyond visual question answering and can also be effective in broader knowledge-intensive domains.

\section{Ethical Considerations}
This work studies multimodal retrieval-augmented generation for knowledge-intensive visual question answering. Our approach may inherit biases, factual errors, and coverage limitations from the underlying models, retrieved knowledge sources, and benchmark datasets, which can lead to misleading or unfair outputs. These risks are particularly relevant for ambiguous, time-sensitive, or long-tail questions. Accordingly, our method is intended for research use, and further validation would be needed before real-world deployment.

We use only publicly available datasets, retrieval sources, and open-source or publicly accessible models. Our work does not involve private data or personally identifiable information. We encourage future research on bias analysis, factuality evaluation, retrieval transparency, and risk mitigation for responsible development of multimodal RAG.

\section{The Use of Large Language Models}
A large language model (LLM) was used only for language editing and LaTeX formatting during the preparation of this manuscript. Its use was limited to improving grammar and clarity and assisting with figure and caption formatting. All scientific ideas, methods, experiments, analyses, and conclusions were produced solely by the authors. All edits were reviewed and verified by the authors.


\end{document}